\documentclass[11pt]{article}
\usepackage{fullpage}

\usepackage[utf8]{inputenc} % allow utf-8 input
\usepackage[T1]{fontenc}    % use 8-bit T1 fonts
\usepackage{hyperref}       % hyperlinks
\usepackage{url}            % simple URL typesetting
\usepackage{booktabs}       % professional-quality tables
\usepackage{amsfonts}       % blackboard math symbols
\usepackage{nicefrac}       % compact symbols for 1/2, etc.
\usepackage{microtype}      % microtypography
\usepackage{tcolorbox}
\usepackage{diagbox}

\usepackage{enumerate}
\usepackage[shortlabels]{enumitem}
\usepackage{algorithmic}
\usepackage{algorithm}
\usepackage{subfigure}
\usepackage{graphicx} % more modern
\usepackage{caption}
\usepackage{amsmath}
\usepackage{amsthm}
\usepackage{amssymb}
\usepackage{tikz}
\usepackage{tablefootnote}
\usepackage{multirow}
\usepackage{enumerate}
\usepackage{color}
\usepackage{xcolor}
\usepackage{natbib}

\usetikzlibrary{arrows}

\allowdisplaybreaks[4]

\usepackage{mathrsfs}

\usepackage{hyperref}
\usepackage{bm,todonotes}

\allowdisplaybreaks

\newtheorem{thm}{Theorem}[section]
\newtheorem{lem}{Lemma}[section]
\newtheorem{cor}{Corollary}[section]
\newtheorem{prop}{Proposition}[section]
\newtheorem{asmp}{Assumption}[section]

\newtheorem{rem}{Remark}[section]

\hypersetup{
	colorlinks=true,
	filecolor=blue,
	citecolor = blue,
	urlcolor=cyan,
}

\usepackage{amsmath,amsfonts,bm}

\def\1{\bm{1}}

\def\eps{{\varepsilon}}

\DeclareMathAlphabet{\mathsfit}{\encodingdefault}{\sfdefault}{m}{sl}
\SetMathAlphabet{\mathsfit}{bold}{\encodingdefault}{\sfdefault}{bx}{n}

\def\sG{{\mathbb{G}}}

\def\sI{{\mathbb{I}}}

\def\sS{{\mathbb{S}}}

\def\sV{{\mathbb{V}}}

\def\B{{\bf B}}

\def\D{{\bf D}}

\def\I{{\bf I}}

\def\s{{\bf s}}

\def\U{{\bf U}}

\def\Xi{\boldsymbol{\xi}}

\def\Phi{\boldsymbol{\phi}}

\def\x{{\boldsymbol{x}}}

\def\s{{\boldsymbol{s}}}

\def\a{\boldsymbol{a}}
\def\sA{\boldsymbol{A}}
\def\sB{\boldsymbol{B}}
\def\sC{\boldsymbol{C}}
\def\sD{\boldsymbol{D}}
\def\sbb{\boldsymbol{b}}

\def\sV{\boldsymbol{V}}
\def\sv{\boldsymbol{v}}
\def\sr{\boldsymbol{r}}
\def\seps{\boldsymbol{\eps}}
\def\sh{\boldsymbol{h}}
\def\sI{\boldsymbol{I}}

\def\sU{\boldsymbol{U}}

\def\sS{\boldsymbol{S}}

\def\sG{\boldsymbol{G}}

\def\U{\boldsymbol{U}}

\def\sdelta{\boldsymbol{\delta}}

\def\Y{{\bf Y}}
\def\y{{\bf y}}

\def\0{{\bf 0}}
\def\1{{\bf 1}}

\def\AM{{\mathcal A}}

\def\DM{{\mathcal D}}

\def\GM{{\mathcal G}}
\def\FM{{\mathcal F}}
\def\IM{{\mathcal I}}

\def\NM{{\mathcal N}}

\def\TM{{\mathcal T}}

\def\RB{{\mathbb R}}
\def\EB{{\mathbb E}}
\def\ZB{{\mathbb Z}}
\def\PB{{\mathbb P}}

\def\bg{\bar{\boldsymbol{g}}}
\def\bx{\bar{\boldsymbol{x}}}
\def\by{\bar{\boldsymbol{y}}}

\def\argmin{\mathop{\rm argmin}}

\newcommand{\tabincell}[2]{\begin{tabular}{@{}#1@{}}#2\end{tabular}}  

\title{
 Statistical Estimation and Inference via Local SGD in Federated Learning
}

\author{
	Xiang Li\thanks{School of Mathematical Sciences, Peking University; email: \texttt{lx10077@pku.edu.cn}. } \\
	\and
	Jiadong Liang \thanks{School of Mathematical Sciences, Peking University; email: \texttt{jdliang@pku.edu.cn}. } \\
	\and
	Xiangyu Chang \thanks{Center for Intelligent Decision-Making and Machine Learning, School of Management, Xi’an Jiaotong University; email: \texttt{xiangyuchang@xjtu.edu.cn}. } \\
	\and
	Zhihua Zhang\thanks{School of Mathematical Sciences, Peking University; email: \texttt{zhzhang@math.pku.edu.cn}. } \\
}
\begin{document}

\maketitle

\begin{abstract}%
Federated Learning (FL) makes a large amount of edge computing devices (e.g., mobile phones)  jointly learn a global model without data sharing. In FL, data are generated in a decentralized manner with high heterogeneity. This paper studies how to perform statistical estimation and inference in the federated setting. 
%We first present a worst-case lower bound to show that the {one-shot average} method is likely to fail on some generalized linear %models due to data heterogeneity. 
We analyze the so-called Local SGD, a multi-round estimation procedure that uses intermittent communication to improve communication efficiency. We first establish a {\it functional central limit theorem} that shows the averaged iterates of Local SGD weakly converge to a rescaled Brownian motion. We next provide two iterative inference methods: the {\it plug-in} and the {\it random scaling}. Random scaling constructs an asymptotically pivotal statistic for inference by using the information along the whole Local SGD path. Both the methods are communication efficient and applicable to online data. Our theoretical and empirical results show that Local SGD simultaneously achieves both statistical efficiency and communication efficiency.
\end{abstract}

\section{Introduction}
%Portable devices like phones and tablets have become the primary computing devices for many users.
%The powerful sensors on these devices (including cameras and microphones), coupled with the fact that they are often carried, mean that %they can access unprecedented amounts of data.
%These data often contain sensitive information such as personal identity information and state of health information, and thus should be %protected from unauthorized access of service providers~\citep{poushter2016smartphone}.
%The challenge arises when limited data access together with memory constraints, communication budget, and computation restrictions makes %the traditional statistical estimation and inference methods no longer applicable~\citep{li2020federated,fan2021modern}.
%To address this issue,~\citet{mcmahan2017communication}  proposed a new distributed computing paradigm called  \textit{Federated %Learning} for collaboratively training a global model from data that remote \textit{clients} hold. The clients can only cooperate with a %central server (e.g., service provider) to train the global model without sharing local datasets.

\textit{Federated Learning} is a novel distributed computing paradigm for collaboratively training a global model from data that remote \textit{clients} hold~\citep{mcmahan2017communication}. The clients can only cooperate with a central server (e.g., service provider) to train the global model without sharing local datasets. Thus, federated learning can protect sensitive information that data often contain, such as personal identity information and state of health information,  from unauthorized access of service providers. The challenge arises  when limited data access together with memory constraints, communication budget, and computation restrictions make the traditional statistical estimation and inference methods~\citep{li2020federated,fan2021modern} no longer applicable in the federated learning scenario.
%In this paper we would like to address this challenge. 

A typical federated learning system considers a pool of $K$ clients, in which the $k$-th client has a local dataset consisting of i.i.d.\ samples  from some unknown distribution $\DM_k$.
The central server faces the following distributed optimization problem: 
\begin{equation}
\label{eq:loss}
\min_{\x} \; f(\x)
=  \sum_{k=1}^{K} p_k f_{k}(\x) 
:= \sum_{k=1}^{K} p_k \EB_{\xi_k \sim \DM_k} f_k(\x; \xi_k),
\end{equation}
where $p_k$ is the weight of the $k$-th client and $f_k(\cdot; \xi_k)$ is the user-specified loss with $\xi_k$ being the generated sample from $\DM_k$.
Due to the decentralized nature of data generation, a discrepancy among local data distributions occurs, i.e., $\{\DM_k\}_{k=1}^K$ are no longer necessarily identical.
In addition, communication is highly restrictive because data with immense volume are scattered across different remote clients.

Many efficient algorithms are proposed to cope with both statistical heterogeneity and expensive communication cost.
Perhaps one of the simplest and most celebrated algorithms for federated learning is \textit{Local SGD}~\citep{stich2018local} (see Algorithm~\ref{alg:local_sgd}).
%Translating Algorithm~\ref{alg:local_sgd} into words, 
Local SGD runs stochastic gradient descent (SGD) independently in parallel on different clients and averages the sequences only once in a while.
Put simple, it learns a shared global model via infrequent communication.
It has been shown to have superior performance in training efficiency and scalability~\citep{lin2018don}, and converge fast in terms of communication~\citep{li2019communication,bayoumi2020tighter,koloskova2020unified,woodworth2020local,woodworth2020minibatch,koloskova2020unified}.  In order to reduce the communication frequency,  Local SGD might also be the best choice. 
%The idea of lowering communication frequency for improving communication efficiency also motivates algorithms for other federated %learning problems, including minimax problems~\citep{mohri2019agnostic,reisizadeh2020robust,deng2021local} and distributed %PCA~\citep{grammenos2019federated,li2021communication}.

 \begin{algorithm}[tb]
 	\caption{Local SGD}
 	\label{alg:local_sgd}
 	\begin{algorithmic}
 		\STATE {\bfseries Input:} functions $\{f_k\}_{k=1}^K$, initial point $\x_0$, step size $\eta_0$, communication set $\IM = \{t_0, t_1, \cdots\}$.
 		\STATE {\bfseries Initialization:} let $\x_{0}^{k} = \x_0$ for all $k$.
 		\FOR{round $m=0$ {\bfseries to} $T-1$}
 		 \FOR{iteration $t=t_{m}+1$ {\bfseries to} $t_{m+1}$}
 		\FOR {each device $k=1$ {\bfseries to} $K$}
 		\STATE  {$\x_t^{k} = \x_{t-1}^{k} - \eta_{m} \nabla f_k(\x_{t-1}^{k}; \xi_{t-1}^{k}).$   \quad \# perform $E_m = t_{m+1}-t_m$ steps of local updates.}
 		\ENDFOR
  		\STATE {The central server aggregates: $\bx_{t_{m+1}} = \sum_{k=1}^K p_k \x_{t_{m+1}}^{k}.$}
  		\STATE {Synchronization: $\x_{t_{m+1}}^k \gets \bx_{t_{m+1}}$ for all $k$.}
 		\ENDFOR
 		 \ENDFOR
 		\STATE {\bfseries Return: $\widehat{\x} = \frac{1}{T} \sum_{m=1}^T\bx_{t_m}$.} 
 	\end{algorithmic}
 \end{algorithm}

From a statistical viewpoint, it is vital to perform statistical inference in federated learning because it helps us infer properties of the underlying data distribution.
However, it is still open how to do that and adapt to the peculiarity of federated learning.
In this paper we would like to address statistical estimation and inference via Local SGD  due to its elegant performance  mentioned earlier  and representativeness in federated learning.

In Local SGD, communication happens at iterations in a prescribed set (denoted $\IM = \{t_0, t_1, t_2, \ldots\}$).
Our goal is to obtain an efficient estimate of $\x^* = \argmin_\x f(\x)$ only through the SGD iterates $\{\x_{t_m}^k\}_{m \in [T], k\in[K]}$, and provide asymptotic confidence intervals for further inference.
Here  $[N]=\{1, 2,\ldots, N\}$ and $\x_{t}^k$ denotes the parameter hosted by the $k$-th client at iteration $t$.
Note that we do not have direct access to $\{\x_{t}^k\}_{k \in [K]}$ if $t \notin \IM$ due to intermittent communication.
It makes the analysis of asymptotic behaviors of Local SGD totally different from that of so-called parallel SGD~\citep{zinkevich2010parallelized}, which alternates between one independent step of SGD in parallel and one synchronization. Clearly, the parallel SGD is equivalent to the single-machine SGD, whose asymptotic convergence has been studied extensively~\citep{blum1954approximation,polyak1992acceleration,anastasiou2019normal,mou2020linear}.

The following questions  emerge:
(i) how one constructs the estimator from Local SGD iterates;
(ii) how local updates (or intermittent communication) affect its asymptotic behavior;
(iii) how one quantifies the variability and randomness of the estimator.
\citet{ruppert1988efficient,polyak1992acceleration} introduced averaged SGD, a simple
modification of SGD where iterates are averaged as the final estimator, and established asymptotic normality via martingale central limit theorem (CLT). 
% This has been recently extended to the non-asymptotic scenario~\citep{anastasiou2019normal,mou2020linear}.
It is known that the averaged SGD estimator obtains the optimal asymptotic variance without any problem-dependent knowledge (e.g., the Hessian at the optima) under certain regularity conditions.
We are motivated to employ the average of Local SGD iterates as the estimator, that is,
\[
\widehat{\x} = \frac{1}{T}\sum_{m=1}^{T} \bx_{t_m} 
\quad \text{where} \quad
\bx_{t_{m}} = \sum_{k=1}^K p_k \x_{t_{m}}^{k}.
\]

The second question has been partially answered in optimization literature~\citep{bayoumi2020tighter,woodworth2020local,woodworth2020minibatch,woodworth2021min}.
It is found local updates slightly slow down the $L_2$ convergence rate of Local SGD, with an additional high-order residual error term dominated by the statistical error term.
It suggests the Local SGD estimator might still have the optimal asymptotic variance even though it has communications with possibly enlarging intermittency.
However, the analysis on its asymptotic behavior is a different story because local updates together with data heterogeneity push local parameters towards different directions before the next-round communication arrives.

As for the last question, we prefer a fully online approach to avoiding repeated large-scale computation. We further can adapt two recent developments in SGD inference.
One is the plug-in method~\citep{chen2020statistical}, which is available when there is an explicit formula for the covariance matrix of the estimator, though having a vast computation cost if the parameter dimension $d$ is too large.
The other borrows insights from time series regression in econometrics~\citep{kiefer2000simple,sun2014let}.
It does not attempt to estimate the asymptotic variance but to construct an asymptotically pivotal statistic by normalizing the estimator with its random transformation.
So it is also known as random scaling~\citep{lee2021fast}.

% \subsection{Contribution}

In this paper we explore possible ways to conduct statistical estimation and inference via Local SGD in FL.
Under common assumptions, we show the proposed estimator $\widehat{\x}$ exactly has the optimal asymptotic variance up to a known scale $\nu (\ge 1)$ which is determined by the sequence $\{E_m\}_m$, where $E_m := t_{m+1}-t_m$ is the length of the $m$-th communication round. {
$\nu$ barely affects the variance optimality because there exist many diverging sequences $\{E_m\}_m$ satisfying $E_m = o(m)$ and $\nu = 1$.
}
In this case, the averaged communication frequency (ACF, i.e., ${T}/{t
_T}$) converges to zero, implying we trade almost all computation for asymptotically zero communication.
Therefore, our estimator simultaneously has statistical efficiency and communication efficiency.

Then, we provide two fully online approaches quantifying the uncertainty of $\widehat{\x}$ with provable guarantees.
One is the plug-in method which estimates the covariance matrix using the iterates from Local SGD.
The other approach constructs an asymptotically pivotal statistic via random scaling without formulating the Hessian matrix.
We establish a functional central limit theorem (FCLT) for the average of Local SGD iterates under mild conditions, which rigorously underpins this approach.
In particular, we develop inequalities to show the non-asymptotic term uniformly vanishes in probability. We believe that the advanced proof technique we developed beyond the current work would be of independent interest.

The remainder of this paper is organized as follows.
% In Section~\ref{sec:failure}, we construct a logistic regression to illustrate the worst-case lower bound of one-shot methods.
In Section~\ref{sec:problem} we formulate our problem and review related work.  
In Section~\ref{sec:estimation} we explore the asymptotic properties for the averaged sequence of Local SGD.
%and collect possible communication intervals $\{E_m\}$.
In Section~\ref{sec:inference} we introduce two online methods (namely the plug-in method and random scaling) to provide asymptotic confidence intervals and perform hypothesis tests.
We illustrate the numerical performance of our methods in synthetic data in Section~\ref{sec:experiments}.
We conclude our article in Section~\ref{sec:conclusion} with a discussion of our results and future research directions.
We defer all the proofs to the appendix.
% and some supplementary results as well discussions

% In this section, establish a functional central limit theorem (FCLT) for the average of Local SGD iterates.

\section{Problem Formulation} 
\label{sec:problem}

In this section, we detail some preliminaries to prepare the readers for our results.
%We then turn to multi-round distributed estimation procedures.
We are concerned with multi-round distributed learning methods.
% This is because one-shot averaging method is likely to fail in the  federated setting as illustrated in Section~\ref{sec:failure}.
At iteration $t$, we use $\x_{t}^k$ to denote the parameter held by the $k$-th client and $\xi_t^k$ the sample it generates according to $\DM_k$.
A typical example of multi-round methods is the parallel stochastic gradient descent (P-SGD)~\citep{zinkevich2010parallelized} that runs
\[
\x_{t+1}^k =\sum_{k=1}^Kp_k\left[ \x_t^k - \eta_t \nabla f_k(\x_t^k; \xi_t^k) \right]
\]
for $k \in [K]$. 
Other variants have been successively  proposed~\citep{jordan2019communication,fan2019communication,chen2021first}.
It is easy to analyze the statistical property of P-SGD due to its equivalence to the single-machine counterpart.
The classical work provides an analysis paradigm for P-SGD, showing it obtains an asymptotically unbiased and efficient estimate~\citep{polyak1992acceleration}.
In particular, with $\bx_t = \sum_{k=1}^Kp_k\x_t^k$, P-SGD achieves the following asymptotic normality with the asymptotic variance satisfying the Cram\'{e}r-Rao lower bound
\[
\sqrt{T}\left(\frac{1}{T}\sum_{t=1}^T\bx_t - \x^*  \right) \overset{d}{\longrightarrow} \NM\left(\0, \; \sG^{-1} \sS \sG^{-\top}  \right),
\]
where $\sG := \nabla^2f(\x^*) = \sum_{k=1}^K p_k \nabla^2 f_k(\x^*)$ is the Hessian at the optima $\x^*$ and $\sS = \EB (\eps(\x^*) \eps(\x^*)^\top) $ is the covariance matrix at it.
Here $\eps(\x^*) = \sum_{k=1}^K p_k \left(\nabla f_k(\x^*; \xi_k) - \nabla f_k(\x^*) \right)$ is the noise of corresponding aggregated gradients.

An obvious drawback of P-SGD is its huge communication because it requires synchronization at each iteration.
By contrast, Local SGD hopes improve the communication efficiency by lowering the communication frequency~\citep{lin2018don,stich2018local,bayoumi2020tighter,woodworth2020local,woodworth2020minibatch}.

\subsection{Local SGD}

%t iteration $t$, we use $\x_{t}^k$ to denote the parameter held by the $k$-th client and $\xi_t^k$ the sample it generates according to %$\DM_k$.
We now turn to  Local SGD  and summarize its details in Algorithm~\ref{alg:local_sgd}.
Put simple, it obtains the solution estimate using the following recursive algorithm
\begin{align}
\label{eq:update-non-linear}
 \x_{t+1}^k =\left\{\begin{array}{ll}
\x_t^k - \eta_t \nabla f_k(\x_t^k; \xi_t^k)           & \text{ if } t+1 \notin \IM, \\
\sum_{k=1}^Kp_k\left[ \x_t^k - \eta_t \nabla f_k(\x_t^k; \xi_t^k)\right]   & \text{ if } t+1  \in \IM,
\end{array}\right. 
\end{align}
where $\eta_t$ is the learning rate, $\xi_t^k$ is an independent realization of $\DM_k$, 
and $\IM$ denotes the set of communication iterations.
At iteration $t$, each client runs always SGD independently in parallel $\x_{t+1}^k  = \x_t^k - \eta_t \nabla f_k(\x_t^k; \xi_t^k)$.
However, when $t+1 \in \IM$, the central server aggregates local parameters $\sum_{k=1}^K p_k \x_{t+1}^k$ and broadcasts it to all clients, which amounts to the following update rule $ \x_{t+1}^k = \sum_{k=1}^Kp_k\left[ \x_t^k - \eta_t \nabla f_k(\x_t^k; \xi_t^k)\right]$.
We remark that our setting also incorporates the finite-sum minimization beyond the stochastic approximation by setting $\DM_k$ as a uniform distribution on a finite set of discrete sample points.

Different choices of $\IM$ lead to different communication efficiency for Local SGD.
% The most evident difference is communication happens only at iterations in $\IM$.
If $\IM = \{0, 1, 2, \cdots \}$, then Local SGD is reduced to P-SGD.
A famous example in practice is constant communication interval~\citep{mcmahan2017communication}, i.e., $\IM = \{0, E, 2E, \cdots \}$ for a predefined integer $E (\ge 1)$, which reduces communication frequency from $1$ to $1/E$.
Local SGD differs from P-SGD if $\IM$ has a general form of $\{t_0, t_1, t_2, \cdots \}$ with some $t_m - t_{m-1} > 1$ where $t_m$ is the $m$-th communication iteration. 
For example, when $t_m< t < t_{m+1}$ for some $m$, $\x_t^k$ is not likely to equal to $\x_t^{k'}$ for $k \neq k'$ due to data heterogeneity, while we always have $\x_t^k = \x_t^{k'}$ for all $k, k'$ for P-SGD.
This difference makes theoretical analysis difficult and different from previous analysis.

For seek of simplicity, we assume $\eta_t$ is a constant when $t_m < t \le t_{m+1}$ and denote it by $\eta_m$ with a little abuse of notation (which has been already adopted in Algorithm~\ref{alg:local_sgd}). In this paper, we study statistical estimation and inference via Local SGD (see Sections~\ref{sec:estimation} and \ref{sec:inference}). 
%First of all, we will show in the following section that the one-shot average is likely to fail in FL due to data heterogeneity,
%which also motivates us to address the multi-round distributed Local SGD. 
First of all, we present some related work, which motivates us to  to address the multi-round distributed Local SGD. 

\subsection{Related Work}

Federated learning enables a large amount of edge computing devices to jointly
learn a global model without data sharing~\citep{kairouz2019advances}.
In the seminal paper~\cite{mcmahan2017communication} proposed \textit{Federated Average} (FedAvg) for FL, which is slightly different from Local SGD that we focus on in this work.
The main difference is that FedAvg randomly samples a small portion of clients at the beginning of each communication round to alleviate the straggler effect caused by massively distributed clients.
When all clients are forced to participate, FedAvg is reduced to Local SGD.
Their theoretical convergence does not vary too much with an additional statistical error incurred when clients participate partially~\citep{li2019convergence}.
There has been a rapidly growing line of work concerning various aspects of FedAvg and its variants recently~\citep{zhao2018federated,sahu2018convergence,nishio2018client,koloskova2020unified,yuan2020federated,yuan2021federated,zheng2021federated}.

In the context of statistical inference, as we know that no works consider the asymptotic properties of Local SGD or FedAvg, letting alone conduct inference.
Most works focus on the optimization properties of Local SGD (or their proposed variants).
~\citet{woodworth2020minibatch,woodworth2020local} gave the state-of-the-art convergence analysis for Local SGD in convex settings, showing its convergence rate is dominated by the statistical error incurred by stochastic approximation of gradients. 
However, it additionally suffers a relatively minor residual error caused by local updates.
As a complementary, our work shows that when the \textit{effective step size} is set to $\gamma_m=E_m\eta_m \propto m^{-\alpha} (\alpha \in (0.5, 1), m\geq 1)$, Local SGD enjoys the optimal asymptotic variance, even though the communication length increases at a sub-linear rate (i.e., $E_m = o(t_m^{1/2})$).
It corresponds to the previous non-asymptotic result~\citep{wang2018cooperative} that shows $E_m$ can be set as large as $O(t_m^{1/2})$ for convergence.
Later, \citet{haddadpour2019local} provided a tighter analysis showing $E_m$ can be set as large as $O(t_m^{2/3})$.
However, they used a smaller learning rate $\gamma_m \propto m^{-1}$ that cannot guarantee asymptotic normality in our theory.
Indeed, the choice of learning rate plays an important role in chasing the non-asymptotic goal of a fast finite-time convergence rate and the asymptotic goal of achieving limiting optimal normality, as noted in~\citet{li2020root} who instead proposed a new SGD variant to achieve both together.
In addition,~\citet{karimireddy2019scaffold,liang2019variance,pathak2020fedsplit,zhang2020fedpd} removed the effect of statistical heterogeneity via control variates or primal-dual techniques.
From our theory, statistical heterogeneity will not affect the asymptotic variance.

Statistical estimation and inference via SGD attracts great attention.
~\cite{ruppert1988efficient,polyak1992acceleration} showed averaging iterates along the SGD trajectory has favorable statistical properties in the asymptotic setting, while~\cite{anastasiou2019normal,mou2020linear} supplemented it with a non-asymptotic analysis.
Many papers recently developed iterative algorithms for constructing asymptotically valid confidence intervals.
~\citet{chen2020statistical} proposed a consistent plug-in estimator.
However, the computation of the Hessian matrix of loss function is not always tractable.
Then,~\citet{chen2020statistical} adapted the non-overlapping batch-means method~\citep{glynn1991estimating} and obtained an offline consistent covariance estimator by using time-increasing batch sizes.
Later on,~\citet{zhu2021online} extended it to a fully online setting via a recursive counterpart using overlapping batches.
In one latest work, ~\citet{lee2021fast} proposed random scaling, which uses nested batches instead. 
But the analysis in their corrected version requires a stronger condition on the gradient noises that should not only be $\alpha$-mixing but also have at least forth-order moment (see their Assumption 2).
The $\alpha$-mixing assumption forces gradient noises to be asymptotic stationary in a fast rate.
By contrast, we provide a valid analysis for random scaling under only $2+\delta$ moment assumptions (see Assumption~\ref{asmp:noise}), which is much weaker and can be of independent interest,
In addition, \citet{fang2018online,fang2019scalable} proposed online bootstrap procedures for the
estimation of confidence intervals via randomly perturbed SGD.
Meanwhile,~\citet{li2018statistical,su2018uncertainty,liang2019statistical} proposed variants of the SGD algorithm to facilitate inference in a non-asymptotic fashion.

\section{Statistical Estimation in Federated Learning}
\label{sec:estimation}

This section aims to provide asymptotic properties for Local SGD.
We start by stating the assumptions needed for the main theoretical results.
These assumptions are quite standard and most of them have been used previously~\citep{polyak1992acceleration,su2018uncertainty,chen2020statistical,li2020root}.

\begin{asmp}[Regularity of the objective]
	\label{asmp:convex}
	For each $k \in [K]$, we assume the objective function $f_k(\cdot)$ is differentiable and strongly convex with parameter $\mu > 0$, i.e., for any $\x, \y$,
\[
f_k(\x) \ge f_k(\y) + \langle \nabla f_k(\y), \x - \y \rangle + \frac{\mu}{2}\|\x-\y\|^2.
\]
In addition, each $f_k(\cdot)$ is $L$-average smooth, i.e.,
\begin{equation}
\label{eq:L-average}
\sqrt{\EB_{\xi_k}\|\nabla f_k(\x; \xi_k) - \nabla f_k(\y; \xi_k)\|^2} \le L \|\x - \y\|
\end{equation}
for some $L >0$.
Finally, the Hessian matrix of the global $f(\cdot)$ exists and is Lipschitz continuous in a neighborhood of the global optimal $\x^*$, i.e., there exist some $\delta_1 > 0$ and  $L' > 0$ such that
\[
\|\nabla^2 f(\x) - \nabla^2 f(\x^*)\| \le L' \|\x - \x^*\|
\]
whenever $\|\x -\x^*\| \le \delta_1$. 
\end{asmp}

Assumption~\ref{asmp:convex} imposes regularity conditions on the objective functions.
It requires the global function $f(\cdot)$ to be $\mu$-strongly convex and $L$-average smooth.
The $L$-average smoothness is stronger than $L$-smoothness because $\|\nabla f_k(\x) - \nabla f_k(\y)\| \le \sqrt{\EB_{\xi_k}\|\nabla f_k(\x; \xi_k) - \nabla f_k(\y; \xi_k)\|^2} \le L \|\x - \y\|$ from Jensen's inequality.
The $L$-average smoothness follows if $\max_{\x}\EB_{\xi_k}\|\nabla^2 f_k(\x; \xi_k)\|^2 < \infty$\footnote{This condition is also made by~\cite{su2018uncertainty} to validate~\eqref{eq:cont-variance}. See Lemma C.1 therein.} which holds for many statistical learning models  such as linear and logistic regression.

Define $\eps_k(\x) = \nabla f_k(\x; \xi_k) - \nabla f_k(\x)$ as the gradient noise at $\nabla f_k(\x)$,  $\sS_k = \EB_{\xi_k} (\eps_k(\x^*)\eps_k(\x^*)^\top)$, and $\eps(\x) = \sum_{k=1}^K p_k \eps_k(\x)$.
Then $\eps_k(\x)$ (as well as $\eps(\x)$) has zero mean and its distribution typically depends on $\x$.
The following assumption regularizes the behavior of each noise $\xi_k$.

\begin{asmp}[Regularized gradient noise]
	\label{asmp:noise}
	We assume the $\xi_k$ on different devices are independent, though they likely have different distributions.
	There exists some $C >0$ such that for each $k \in [K]$, 
	\begin{equation}
	 \label{eq:cont-variance}
	 \left\| \EB_{\xi_k} (\eps_k(\x) \eps_k(\x)^\top) - \sS_k \right\| \le C \left[ \| \x - \x^*\| +\| \x - \x^*\|^2 \right]. 
	\end{equation}
	Moreover, we assume there exists a constant $\delta_2 > 0$ such that
	\[
	\sup_{\| \x - \x^*\| \le \delta_2 } \; \EB\|\eps(\x)\|^{2+\delta_2} < \infty.
	\]
\end{asmp}

Assumption~\ref{asmp:noise} first requites the $\xi_k$ are mutually independent. 
%Let $\sS_k = \EB_{\xi_k} \eps_k(\x^*)\eps_k(\x^*)^\top$. 
Note that $\sS = \sum_{k=1}^K p_k^2 \sS_k$ is the Hessian at the optimum $\x^*$ because $\sS =\sum_{k=1}^K p_k^2 \EB_{\xi_k } (\eps_k(\x^*)\eps_k(\x^*)^\top) = \EB_{\xi} (\eps(\x^*)\eps(\x^*)^\top )$ from the independence assumption.
It then forces the difference between covariance matrices $\EB_{\xi_k} (\eps_k(\x) \eps_k(\x)^\top)$ and $\sS_k$ controlled by $\|\x - \x^*\|$.
It implies $\left\| \EB_{\xi} (\eps(\x) \eps(\x)^\top) - \sS \right\| \le C' \left[ \| \x {-} \x^*\| +\| \x{-} \x^*\|^2 \right]$.
Finally, the imposed uniformly finite $(2+\delta_2)$ moment of $\eps(\cdot)$ around the optimum $\x^*$ establishes the Lindeberg-Feller condition for martingales.

\begin{asmp}[Slowly decaying effective step sizes]
\label{asmp:gamma}
 Define $\gamma_m = E_m \eta_m$ as the effective step size, and assume it is non-increasing in $m$ and satisfies \emph{(i)} $\sum_{m=1}^\infty \gamma_m^2 < \infty$; \emph{(ii)} $\sum_{m=1}^\infty \gamma_m = \infty$; and \emph{(iii)} $\frac{\gamma_m - \gamma_{m+1}}{\gamma_m} = o(\gamma_m)$.
\end{asmp}

In our analysis, $\gamma_m = E_m \eta_m$ serves as the \textit{effective step size}.
Indeed, the previous analysis of~\citet{li2019convergence} shows that the effect of $E_m$ steps of local updates with step-size $\eta_t$ is similar to one-step update with a larger step-size $E_m \eta_m$. 
It implies that it is the multiplication of $E_m$ and $\eta_m$, rather than either of them alone effecting the convergence.
A typical example satisfying the assumption is $\gamma_m = \gamma m^{-\alpha}$ with $\alpha \in (0.5,1)$, which is also frequently used in previous works~\citep{polyak1992acceleration,chen2020statistical,su2018uncertainty}.
Because we impose restriction to $\{E_m\}$ latter, in practice, we can first determine the sequence of $\{E_m\}$ and then set $\eta_m = \gamma_m/E_m$ to meet the requirement of $\{\gamma_m\}$.

\begin{asmp}[Slowly increasing communication intervals]
\label{asmp:E}
The sequence $\{E_m\}$ satisfies 
\begin{enumerate}%[(i)]
	\item[\emph{(i)}] $\{E_m\}$ is either uniformly bounded or non-decreasing;
	\item[\emph{(ii)}] There exists some $\delta_3 > 0$ such that $\limsup\limits_{T \to \infty} \frac{1}{T^2} (\sum_{m=0}^{T-1} E_m^{1+\delta_3})(\sum_{m=0}^{T-1} E_m^{-(1+\delta_3)}) < \infty$;
	\item[\emph{(iii)}] 
	\[
	\lim\limits_{T \to \infty} \frac{1}{T^2} (\sum_{m=0}^{T-1} E_m)(\sum_{m=0}^{T-1} E_m^{-1}) = \nu (\nu \ge 1);
	\]
	\item[\emph{(iv)}] $\lim\limits_{T \to \infty} \frac{\sqrt{t_T}}{T} \cdot \left( \sum\limits_{m=0}^{T}\gamma_m\right) = 0$ and $\lim\limits_{T \to \infty} \frac{\sqrt{t_T}}{T } \frac{1}{\sqrt{\gamma_T}}= 0$ where $t_T = \sum_{m=0}^{T-1} E_m$.
\end{enumerate}
\end{asmp}

Assumption~\ref{asmp:E} restricts the growth of $\{E_m\}$.
Intuitively, if $E_m$ increases too fast, each $\x_{t}^k$ might converge to their local minimizer $\x_k^*$ rapidly before the next communication.
Therefore, their average $\bx_t$ is asymptotically biased for $\x^*$ with  the bias $\sum_{k=1}^K p_k \x_k^* - \x^*$, which is unlikely  zero in FL.
%Mathematically speaking, 
Because $\sum\limits_{m=0}^{T-1} \gamma_m \ge \gamma_0$, we have $\sqrt{t_T}/T = {\sqrt{\sum_{m=0}^{T-1} E_m}}/{T} \to 0$ from $\text{(iv)}$.
This, combined with $\text{(iii)}$, implies $\sum_{m=0}^{T} E_m^{-1} \to \infty$.
It forbids $\{E_m\}$ from growing too fast.
In practice, we can choose $E_m \sim \ln m$, $E_m \sim \ln\ln m$ or $E_m \sim m^{\beta}$ with $\beta \in (0, 1)$, all of them satisfying $\text{(ii)}$ and $\text{(iii)}$.
If $\gamma_m \sim m^{-\alpha}$ with $\alpha \in (0.5, 1)$, all the choices of $E_m$ above satisfy $\text{(iv)}$.

The following proposition provides another way to check $\text{(ii)}$ and $\text{(iii)}$ in Assumption~\ref{asmp:E} via investigating the relative difference of $E_m$ and $E_{m-1}$.

\begin{prop}\label{prop:limit-Em}
Assume $\{E_m\}$ is non-decreasing.
If $\limsup\limits_{m\to \infty} m (1-\frac{E_{m-1}}{E_m}) < 1$, then $(ii)$ in Assumption~\ref{asmp:E} holds for some $\delta_3 > 0$.
Furthermore, if $\lim\limits_{m \to \infty} m (1-\frac{E_{m-1}}{E_m})$ exists (denoted  $\rho$), once $\rho < 1$, then $(iii)$ in Assumption~\ref{asmp:E} holds with
\[
\nu = \frac{1}{1-\rho^2}.
\]
\end{prop}

According to the aforementioned regularity assumptions, the following asymptotic normality property of the averaged iterates generated by Local SGD is investigated in Theorem~\ref{thm:clt}.
\begin{thm}[Asymptotic Normality]
	\label{thm:clt}
	Let Assumptions~\ref{asmp:convex},~\ref{asmp:noise} and~\ref{asmp:gamma} hold. Then $\bx_{t_m}$ converges to $\x^*$ not only almost surely but also in $L_2$ convergence sense with rate $$\EB\|\bx_{t_m}-\x^*\|^2 \lesssim \gamma_m.$$
	Moreover, if Assumption~\ref{asmp:E} holds additionally, the asymptotic normality follows
	\[
	\sqrt{t_T}  \left( \frac{1}{T}\sum_{m=1}^T \Bar{\x}_{t_m} -\x^*\right) \overset{d}{\longrightarrow} \NM\left(\0, \; \nu \sG^{-1} \sS \sG^{-\top}  \right),
	\]
	where $t_T = \sum_{m=0}^{T-1}E_m$, $\bx_{t_m} = \sum_{k=1}^Kp_k \x_{t_m}^k$, $\sG = \sum_{k=1}^K p_k \nabla^2 f_k(\x^*)$ is the Hessian matrix at the optima $\x^*$, and $\sS$ is the covariance matrix of aggregated gradient noise.
\end{thm}

\begin{table}[t!]
	\caption{Statistical efficiency and communication efficiency under different choices of $E_m, \gamma_m$ and $\eta_m$. The statistical efficiency is measured by $\nu$, while the communication efficiency is measured by averaged communication frequency (ACF), i.e., ${T}/{\sum_{m=0}^{T-1} E_m}$.} 
	\centering 
	\begin{tabular}{c| c |c| c| c| c} 
		\hline\hline 
		Case & $E_m (\ge 1)$ & $\gamma_m $ & $\eta_m$ & $\nu (\ge 1)$ & ACF\\
		\hline
		Base  & $1$ & 
		\multirow{6}{*}{\tabincell{c}{$\gamma m^{-\alpha} $\\$\alpha \in (0.5, 1)$}}
		   &  ${\gamma} m^{-\alpha}$  &1 & $1$ \\
		1  & $E$ &     &  ${\gamma} m^{-\alpha}/{E}$  &1 & $E^{-1}$ \\
		2  & any $E_m\le E$ &      &  ${\gamma} m^{-\alpha}/{E_m}$  &1 & $[E^{-1}, 1]$ \\
		3 & $E\ln^\beta m \ (\beta > 0)$  &   &  $\gamma m^{-\alpha}/(E \ln^\beta m)$  &  1 & $E^{-1} \ln^{-\beta} T$  \\
		4 & $E\ln^\beta\ln m \ (\beta > 0)$  &  &  $\gamma m^{-\alpha}/(E \ln^\beta \ln m)$  &  1 & $E^{-1} \ln^{-\beta} \ln T$  \\
		5 & $E m^{\beta} \ (\beta \in (0, 1))$  &   &  ${\gamma} m^{-(\alpha+\beta)}/{E}$  &  $(1-\beta^2)^{-1}$ & $(1+\beta) E^{-1} T^{-\beta} $\\
		\hline 
		\hline 
	\end{tabular}
	\label{table:1} 
\end{table}

Theorem~\ref{thm:clt} shows that the averaged sequence generated by Local SGD has an asymptotic normal distribution with the asymptotic variance depending on how communication happens (i.e., the sequence $\{E_m\}$) and the problem parameters (i.e., $\sS$ and $\sG$).
For one thing, the effect of data heterogeneity doesn't show up in the asymptotic normality.
The asymptotic variance as well as $L_2$ convergence rate is the same with that of P-SGD.
This is because the residual error caused by data heterogeneity typically has relatively low order than the statistical error incurred by stochastic gradients~\citep{woodworth2020minibatch,woodworth2020local}.
With the choice of $\gamma_m$, the residual error vanishes much faster and then seems to disappear.
For another thing, it is quite interesting that the whole optimization process affects the asymptotic variance.
At the worst case, the way how communication frequency is determined only enlarges the asymptotic variance by a known scale $\nu(\ge 1)$.
If $E_m \equiv 1$ for all $m$ (which implies no local update is called), $\nu = 1$ and the result is identical to the typical single-machine central limit theorem (CLT) for SGD~\citep{polyak1992acceleration}.
When $E_m$ varies, it is still possible to get communication saved and the asymptotic variance unchanged (i.e., $\nu = 1$) simultaneously (see Table~\ref{table:1}).
If $E_m$ is uniformly bounded or grows in a rate slower than $E \ln^\beta m (\beta > 0)$, we maintain $\nu = 1$ and obtain a smaller average communication frequency (ACF).
In the latter case, the ACF is asymptotic zero, which implies that we trade almost all computation for nearly zero communication without any sacrifice for statistical efficiency.
However, if $E_m$ grows like $E m^{\beta} \ (\beta \in (0, 1))$, though its ACF decays much more rapidly than that of $E \ln^\beta m$, the asymptotic variance is increased by a factor of $\nu=(1-\beta^2)^{-1}$. 
It depicts a trade-off between communication efficiency and statistical efficiency when $E_m$ grows too fast.
Finally, $E_m$ could not grows like $E m^{\beta} \ (\beta > 1)$ or even exponentially fast, because this will violate the requirement $\sum_{m=0}^{T-1} E_m^{-1} \to \infty$ that is inherent from Assumption~\ref{asmp:E}.

\section{Statistical Inference in Federated Learning}
\label{sec:inference}

We now conduct statistical inference via Local SGD in the FL setting.
As argued in the introduction, the central server only has access to $\{\x_{t}^k\}_{k \in [K]}$ when $t \in \IM$.
In terms of the established CLT (Theorem \ref{thm:clt}), the average of $\{\bx_{t_m}\}_{m\in[T]}$ achieves an asymptotic normality.
Thus it is natural to use $\{\bx_{t_m}\}_{m\in[T]}$ as the main iterate to construct asymptotically valid confidence intervals.
We will refer to $\{\bx_{t_m}\}_{m\in[T]}$ as the \textit{path of Local SGD}.

In this section, we assume the data are generated locally in a fully online fashion because it not only can be reduced to the finite-sample setting via bootstrapping, but also covers many realistic FL settings where data are generated sequentially, typical examples including the records of web search, online shopping, and bank credits.
In particular, we propose two inference methods depending on whether the second order information of the loss function is available.
One is the plug-in method that uses the Hessian information directly and the other is the random scaling method that uses only the information among the path of Local SGD.

%  $\widehat{\sG}^{-1}\widehat{\sS}\widehat{\sG}^{-\top}$. 
% From a completely different perspective, the random scaling method uses the invariant principle of parameter $\x_t$ to derive a pivotal statistics for inference.

\subsection{The Plug-in Method}
The plug-in method first estimates $\sG$ and $\sS$ by $\widehat{\sG}$ and $\widehat{\sS}$, respectively, and obtains the estimator of the covariance matrix with $\widehat{\sG}^{-1}\widehat{\sS}\widehat{\sG}^{-\top}$.
The key  is to obtain consistent estimators $\widehat{\sG}$ and $\widehat{\sS}$.
An intuitive way to construct $\widehat{\sG}$ and $\widehat{\sS}$ is to use the sample estimate as follows
\begin{eqnarray*}
\widehat{\sG}_T &:= & \frac{1}{T}\sum_{m=1}^{T}\sum\limits_{k=1}^K p_k \nabla^2f_k(\bx_{t_{m}};\xi_{t_{m}}^k), \\
\widehat{\sS}_T &:= & \frac{1}{T}\sum_{m=1}^{T}\left(\sum_{k=1}^Kp_k\nabla f_k(\bx_{t_m};\xi_{t_m}^k) \right) \left(\sum_{k=1}^Kp_k\nabla f_k(\bx_{t_m};\xi_{t_m}^k)\right)^\top
\end{eqnarray*}
as long as  each $\nabla^2 f_k(\bx_{t_m}; \xi_{t_m}^k)$ is available.
It is worth noting that with $\bx_{t_m}$, as well as each local Hessian and gradient evaluated at it, communicated to the central server, we can update $\widehat{\sG}_{m-1}$ to $\widehat{\sG}_{m}$ and $\widehat{\sS}_{m-1}$ to $\widehat{\sS}_{m}$.
Therefore, they can be computed in an online manner without the need of storing
all the data.

\begin{asmp}
\label{asmp:H}
There are some constants $L{''} > 0$ such that for any $k \in [K]$,
\begin{gather*}
\EB_{\xi_k}\|\nabla^2f_k(\x; \xi_k) - \nabla^2 f_k(\x^*; \xi_k)\| \le L{''} \|\x -\x^*\|.
\end{gather*}
\end{asmp}

Following~\cite{chen2020statistical},
we make Assumption~\ref{asmp:H}, which slightly strengthens the Hessian smoothness assumption in Assumption~\ref{asmp:convex}. 
%It is also used by.
%With the additional assumption, 
Accordingly, 
we establish the consistency of the
sample estimate $\widehat{\sG}_T$ and $\widehat{\sS}_T$ in the following theorem.

\begin{thm}
\label{thm:G-and-S}
Under Assumptions~\ref{asmp:convex},~\ref{asmp:noise},~\ref{asmp:gamma} and~\ref{asmp:H}, $\widehat{\sG}_T$ and $\widehat{\sS}_T$ converge to $\sG$ and $\sS$ in probability  as $T \to \infty$.
As a result of Slutsky's theorem, $\widehat{\sG}_T^{-1}\widehat{\sS}_T\widehat{\sG}_T^{-\top}$ is consistent to $\sG^{-1}\sS\sG^{-\top}$.
\end{thm}

Theorem~\ref{thm:G-and-S} implies that $(\sG^{-1}\sS\sG^{-\top})_{jj}$ can be estimated by $\widehat{\sigma}_{T,j}^2 = (\widehat{\sG}_T^{-1}\widehat{\sS}_T\widehat{\sG}_T^{-\top})_{jj}$ for the construction of confidence intervals.
Denoting $\by_T = \frac{1}{T} \sum_{m=1}^T \bx_{t_m}$  and $\by_{T, j}$ its $j$-th coordinate,
we have the following corollary which shows that $\by_{T, j} \pm z_{\frac{\alpha}{2}} \sqrt{\frac{\widehat{\nu}_T}{t_T}} \widehat{\sigma}_{T,j}$ constructs an asymptotic exact confidence interval for the $j$-th coordinate of $\x^*$.
Here $\widehat{\nu}_T$ is any sequence converging to $\nu$.

\begin{cor}
Under the assumption of Theorem~\ref{thm:G-and-S}, 
\[
\PB\left(  
\by_{T, j} - z_{\frac{\alpha}{2}}\sqrt{\frac{\widehat{\nu}_T}{t_T}} \widehat{\sigma}_{T,j}
\le \x_j^* \le 
\by_{T, j} + z_{\frac{\alpha}{2}}\sqrt{\frac{\widehat{\nu}_T}{t_T}} \widehat{\sigma}_{T,j}
\right)
\to 1-\alpha,
\]
where $\widehat{\nu}_T \to \nu$ and $z_{\frac{\alpha}{2}}$ is $(1-\alpha/2)$-quantile of the standard normal distribution.
\end{cor}

We remark that using an estimate $\widehat{\nu}_T$ instead of the true value $\nu$ for inference is for the purpose of practice.
We find in experiments that directly using the true value $\nu$ often results in an unstable confidence interval due to slow convergence of $\text{(iii)}$ in Assumption~\ref{asmp:E}.
As a remedy, we use an estimate $\widehat{\nu}_T = \frac{1}{T^2}(\sum_{m=1}^T E_m)(\sum_{m=1}^T E_m^{-1})$ which performs better and more stable.

The plug-in method typically works well in practice due to its simplicity and well-established theoretical guarantees.
However, it has some drawbacks.
The most obvious one is the requirement of the Hessian information, which is not always accessible.
Besides, the formulation and sharing of each $\nabla^2 f_k(\bx_{t_m}; \xi_{t_m}^k)$ requires at least $O(d^2)$ memory and communication cost.
Furthermore, it may be computationally expensive when $d$ is large because it involves matrix inversion with computation complexity $O(d^3)$.
Finally, the inverse operation is unstable empirically. 
In practice, we need to set the round $T$ sufficiently large to avoid singularity and ensure stable estimation.
The estimator introduced in the next subsection provides a fully online approach, which is cheap in memory, computation, and communication.

\subsection{Random Scaling}
Random scaling does not attempt to estimate the asymptotic variance, but studentize $\by_T = \frac{1}{T} \sum_{m=1}^T \bx_{t_m}$ with a matrix constructed using iterates along the Local SGD path.
In this way, an asymptotically pivotal statistic, though not asymptotically normal, can be obtained.
To clarify the method, we should first figure out the asymptotic behavior of the whole Local SGD path rather than its simple average $\by_T$.
In particular, we have the following functional central limit theorem that shows the standardized partial-sum process converges in distribution to a rescaled Brownian motion.

\begin{thm}[Functional CLT]
	\label{thm:fclt}
	Let Assumptions~\ref{asmp:convex},~\ref{asmp:noise},~\ref{asmp:gamma} and~\ref{asmp:E} hold, and define
\[
    h(r, T) = \max\left\{ n \in \ZB, n > 0 \bigg|  r \sum_{m=1}^T \frac{1}{E_m}  \ge  \sum_{m=1}^n \frac{1}{E_m}  \right\}
    \quad \text{for} \quad r \in (0, 1].
	\]
	As $T \to \infty$, the following random function weakly converges to a scaled Brownian motion, i.e.,
	\[
	\phi_T(r) :=
	\frac{\sqrt{t_T}}{T}  \sum_{m=1}^{h(r, T)} \left(\bx_{t_m} -\x^*\right) \Rightarrow  \sqrt{\nu}  \sG^{-1}\sS^{1/2} \B_d(r) 
	\]
	where  $t_T = \sum_{m=0}^{T-1}E_m$, $\bx_{t_m} = \sum_{k=1}^Kp_k \x_{t_m}^k$, and $\B_d(\cdot)$ is the $d$-dimensional standard Brownian motion. 
\end{thm}

Theorem~\ref{thm:fclt} has many implications.
First, the result is stronger than Theorem~\ref{thm:clt} though under the same assumptions.
By applying the continuous mapping theorem to Theorem~\ref{thm:fclt} with $\psi: C[0, 1] \mapsto \psi(1)$, we directly prove Theorem~\ref{thm:clt}.
Hence, we only give the proof of Theorem~\ref{thm:fclt} in Appendix~\ref{append:main-proof}.
Second, the sequence $\{E_m\}$ makes a difference via the time scale $h(r, T)$, which extends previous FCLT results on SGD.
For example, if $E_m \equiv E$, then $\nu=1, t_T = ET$ and $h(r, T) = \lfloor rT \rfloor$, the result turning to be 
\[
\frac{1}{\sqrt{T}}\sum_{m=1}^{ 
\lfloor rT \rfloor} \left(\bx_{t_m} -\x^*\right) \Rightarrow  \sqrt{\frac{1}{E}}  \sG^{-1}\sS^{1/2} \B_d(r).
\]
When $E = 1$, it reduces to the single-machine result that is recently obtained by~\citet{lee2021fast}.
Once $E > 1$, an interesting observation is that local updates reduce the scale of the Brown motion.
As an extreme case, the scale vanishes and the Brown motion degenerates when $E = \infty$.
It makes sense because when $E=\infty$, $\x_{t_m}^k \equiv \x_k^*$ and $\bx_{t_m} \equiv \sum_{k=1}^Kp_k\x_{t_m}^k$, the process degenerates.
Beyond constant $E_m \equiv E$, Theorem~\ref{thm:fclt} also embraces mildly increasing $\{E_m\}$ (see Table~\ref{table:1}).

With Theorem~\ref{thm:fclt}, we are ready to describe the inference method.
Define $r_0 = 0$ and
\[
r_m =  \frac{\sum_{n=1}^m \frac{1}{E_n}}{\sum_{n=1}^T \frac{1}{E_n}}, \; \mbox{ for } m \ge 1.
\]
The choice of $r_m$ satisfies that $\phi_T(r_m) = \frac{\sqrt{t_T}}{T}\sum_{n=1}^m (\bx_{t_n}-\x^*)$.
Note that $\phi_T(1) = \frac{\sqrt{t_T}}{T}\sum_{n=1}^T (\bx_{t_n}-\x^*) = \sqrt{t_T}(\by_T-\x^*)$.
Hence, $\phi_T(r_m)-\frac{m}{T}\phi_T(1) = \frac{\sqrt{t_T}}{T}\sum_{n=1}^m (\bx_{t_n}-m\by_T)$ cancels the dependence on $\x^*$.
To remove the dependence on the unknown scale $\sG^{-1}\sS^{1/2}$, we studentize $\phi_T\left(1\right)$ via
\[
\Pi_T = \sum_{m=1}^T\left(\phi_T(r_m)-\frac{m}{T}\phi_T(1)\right)\left(\phi_T(r_m)-\frac{m}{T}\phi_T\left(1\right)\right)^\top(r_m - r_{m-1}).
\]
\begin{cor}
\label{cor:pivotal}
Under the same assumptions of Theorem~\ref{thm:fclt} and assuming $ g(r_m)\asymp \frac{m}{T}$ for some continuous function $g$ on $[0, 1]$ , we  have that
\[
\phi_T\left(1\right)^\top\Pi_T^{-1}\phi_T\left(1\right)
\overset{d}{\to}
\B_d(1)^\top
\left[\int_{0}^1 \left(\B_d(r) {-} g(r) \B_d(1) \right)\left(\B_d(r) {-} g(r) \B_d(1) \right)^\top d r\right]^{-1}\B_d(1).
\]
\end{cor}
This corollary follows immediately from Theorem~\ref{thm:fclt} and the continuous mapping theorem.
It implies $\phi_T\left(1\right)^\top\Pi_T^{-1}\phi_T\left(1\right)$ is asymptotically pivotal and thus can be used to construct valid asymptotic confidence intervals.
Up to a constant factor, studentizing $\phi_T\left(1\right)$ via $\Pi_T$ is equivalent to studentizing $\by_T = \frac{1}{T} \sum_{m=1}^T \bx_{t_m}$ via $\widehat{\sV}_T$ where
\[
\widehat{\sV}_T 
= \frac{1}{T^2\sum_{m=1}^T \frac{1}{E_m}}\sum_{m=1}^T\frac{1}{E_m}\left(\sum_{n=1}^m\bx_{t_n} - m\by_T\right)\left(\sum_{n=1}^m\bx_{t_n} - m\by_T\right)^\top.
\]
$\widehat{\sV}_T$ can be updated in an online manner.
To state its online updating rule, recall that $\by_m = \frac{1}{m} \sum_{n=1}^m \bx_{t_n}$ and note that
\begin{align*}
\widehat{\sV}_T 
&= \frac{1}{T^2\sum_{m=1}^T \frac{1}{E_m}}\sum_{m=1}^T\frac{m^2}{E_m}\left(\by_m -\by_T\right)\left(\by_m - \by_T\right)^\top\\
&= \frac{1}{T^2\sum_{m=1}^T\frac{1}{E_m}} \left[  
\sum_{m=1}^T\frac{m^2}{E_m}\by_m\by_m^\top - \sum_{m=1}^T\frac{m^2}{E_m}\by_T \by_m^\top -  \sum_{m=1}^T \frac{m^2}{E_m} \by_m\by_T^\top  + \sum_{m=1}^T\frac{m^2}{E_m} \by_T\by_T^\top\right].
\end{align*}
Hence, to update $\widehat{\sV}_{m-1}$ to $\widehat{\sV}_{m}$ when a new observation $\bx_{t_m}$ is available, we only need to keep the following quantities, namely $s_{m-1} = \sum_{n=1}^{m-1} \frac{1}{E_n}$, $q_{m-1} = \sum_{n=1}^{m-1} \frac{n^2}{E_n}$, $\by_{m-1} = \frac{1}{m-1} \sum_{n=1}^{m-1} \bx_{t_n}$, 
\[
\sA_{m-1} = \sum_{n=1}^{m-1}\frac{n^2}{E_n}\by_n\by_n^\top 
\quad \text{and} \quad
\sbb_{m-1} = \sum_{n=1}^{m-1}\frac{n^2}{E_n}\by_n,
\]
all of which can be updated in  online. 
The formal formulation is presented in Algorithm~\ref{alg:local_sgd_ref}.

 \begin{algorithm}[tb]
 	\caption{Online Inference with Local SGD via Random Scaling}
 	\label{alg:local_sgd_ref}
 	\begin{algorithmic}
 		\STATE {\bfseries Input:} functions $\{f_k\}_{k=1}^n$, initial point $\x_0$, step size $\eta_t$, communication set $\IM = \{t_0, t_1, \cdots\}$.
 		\STATE {\bfseries Initialization:} set $\x_{0}^{(k)} = \x_0$ for all $k$, let $\sA_0=\0$ and $\sbb_0=\0$ and $s_0 =q_0= 0$.
 		\FOR{$m=1$ {\bfseries to} $T$}
 		\STATE {Obtain the synchronized variable from Local SGD: $\bx_{t_m} =\sum_{k=1}^K p_k \x_{t_m}^{k}.$ }
 		\STATE {$\by_m = \frac{m-1}{m}\by_{m-1}+ \frac{1}{m}\bx_{t_m},$  }
 		\STATE {$\sA_m = \sA_{m-1} + \frac{m^2}{E_m}\by_m\by_m^\top,$ }
 		\STATE {$\sbb_m = \sbb_{m-1} + \frac{m^2}{E_m} \by_m,$}
 		\STATE {$s_m = s_{m-1} + \frac{1}{E_m},$}
 		\STATE {$q_m = q_{m-1} + \frac{m^2}{E_m}.$}
 		\STATE Obtain $\widehat{\sV}_m$ by
 		\[
 		\widehat{\sV}_m = \frac{1}{m^2s_m} \left( \sA_m - \by_m \sbb_m^\top - \sbb_m \by_m^\top + q_m \by_m \by_m^\top \right).
 		\]
 		 \STATE {\bfseries Return: $\by_m$ and $\widehat{\sV}_m$.}
 		 \ENDFOR
 	\end{algorithmic}
 \end{algorithm}
 
Once $\by_T$ and $\widehat{\sV}_T$ are obtained, it is straightforward to carry out inference.
For example, we construct the $(1{-}\alpha)$ asymptotic confidence interval for the $j$-th element $\x_j^*$ of $\x^*$ as follows
\begin{cor}
Under the same conditions of Corollary~\ref{cor:pivotal}, we have that
\[
\PB\left( \left[ \by_{T,j} - q_{\frac{\alpha}{2},g} \sqrt{\widehat{\sV}_{T,jj}}
\le \x_j^* \le
\by_{T,j} + q_{\frac{\alpha}{2}, g} \sqrt{\widehat{\sV}_{T,jj}}
\right] \right) \to 1-\alpha,
\]
where $q_{\frac{\alpha}{2}, g}$ is $(1-\alpha/2)$-quantile of the following random variable 
\begin{equation}
\label{eq:t}
{B_1(1)}\bigg/{\left(\int_0^1 (B_1(r) - g(r) B_1(1))^2 dr\right)^{1/2}}
\end{equation}
with $B_1(\cdot)$ a one-dimensional standard Brownian motion.
\end{cor}

 \begin{table}[t!]
	\caption{Asymptotic critic values $q_{\alpha, \beta}$ of $t^*(\beta)$ defined by
	$q_{\alpha, \beta} = \min\{t:\PB(t^*(\beta) \le t) \ge 1-\alpha\}$.
	} 
	\centering 
	\begin{tabular}{c|ccccccccc} 
	\toprule
    \diagbox{$\beta$}{$1-\alpha$}& $1\%$ & $2.5\%$ & $5\%$ & $10\%$ & $50\%$ & $90\%$ & $95\%$ & $97.5\%$ &$99\%$ \\
    \midrule		
    0 & -8.634&-6.753&-5.324&-3.877& 0.000 & 3.877 & 5.324& 6.753 & 8.634\\
    $1/3$ &  -8.0945 & -6.339 & -5.048& -3.712&  0.000&3.712 & 5.048 & 6.339& 8.0945\\
    $1/2$ &  -7.386 &  -5.851& -4.621& -3.446& 0.000 &3.446 & 4.621 & 5.851 & 7.386\\
    $2/3$ &  -6.292 &  -4.993 &-4.012 &-3.027& 0.000& 3.027 & 4.012 & 4.993 & 6.292\\
		\bottomrule
	\end{tabular}
	\label{table:2} 
\end{table}

The remaining issue is about the specific form of $g$ and the computation of $q_{\alpha, g}$.
$g$ actually depends on the growth of $\{E_m\}$.
Direct computation reveals that $r_m \asymp \left( \frac{m}{T}\right)^{1-\beta}$ if $E_m \asymp m^{\beta}$ and  $r_m \asymp \frac{m}{T}$ if $E_m \asymp \ln^\beta(m)$.
Hence, we are motivated to consider the following family of $g$: $g_{\beta}(r) = r^{\frac{1}{1-\beta}}$ indexed by $\beta \in [0, 1)$.

With this $g_{\beta}(\cdot)$, we denote the random variable given in~\eqref{eq:t} by $t^*(\beta)$ and the corresponding critical value by $q_{\alpha, \beta} := \min\{t:\PB(t^*(\beta) \le t) \ge 1-\alpha\}$.
The limiting distribution $t^*(\beta)$ is mixed normal and symmetric around zero.
For easy reference, critical values of $t^*(\beta)$ are computed via simulations and listed in Table~\ref{table:2}.
In particular, the Brownian motion $B_1(\cdot)$ is approximated by normalized sums of i.i.d.\ $\NM(0, 1)$ pseudo random deviates using 1,000 steps and 50,000 replications. 
We then smooth the 50,000 realizations by standard Gaussian-kernels techniques with the bandwidth selected according to Scott's rule~\citep{scott2015multivariate}.
Kernel density estimation is a way to estimate the probability density function of a random variable in a non-parametric way.
Because we smooth the data, our critical values of the case $\beta=0$ are slightly different from previous computations by~\citet{kiefer2000simple}.
In particular, when $1-\alpha=97.5\%$ and $\beta = 0$, our critical value $6.753$ is smaller than previous $6.811$, which shrinks the length of our confidence intervals.

\begin{figure}[t!]
	\centering
	\includegraphics[width=\textwidth]{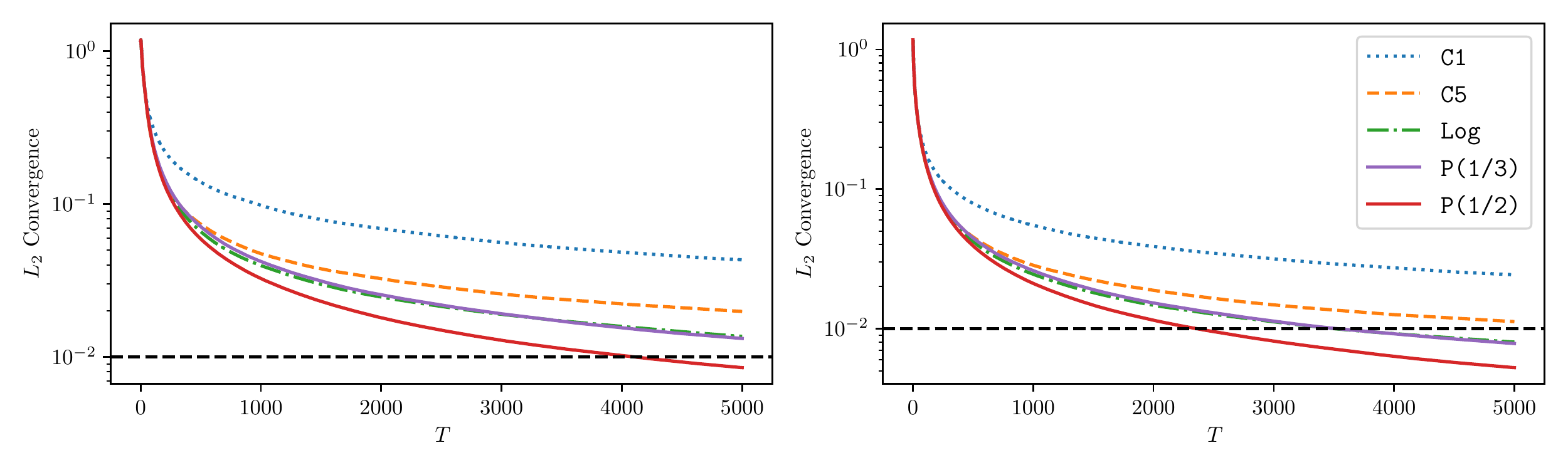}
	\caption{$L_2$ convergence $\|\by_T-\x^*\|$ in terms of communication $T$. 
	Left: Results of linear regression. 
	Right: Results of logistic regression.
	Black dashed line denotes the nominal coverage rate of 95\%.}
	\label{fig:convergence}
\end{figure}
\begin{figure}[t!]
	\centering
	\includegraphics[width=\textwidth]{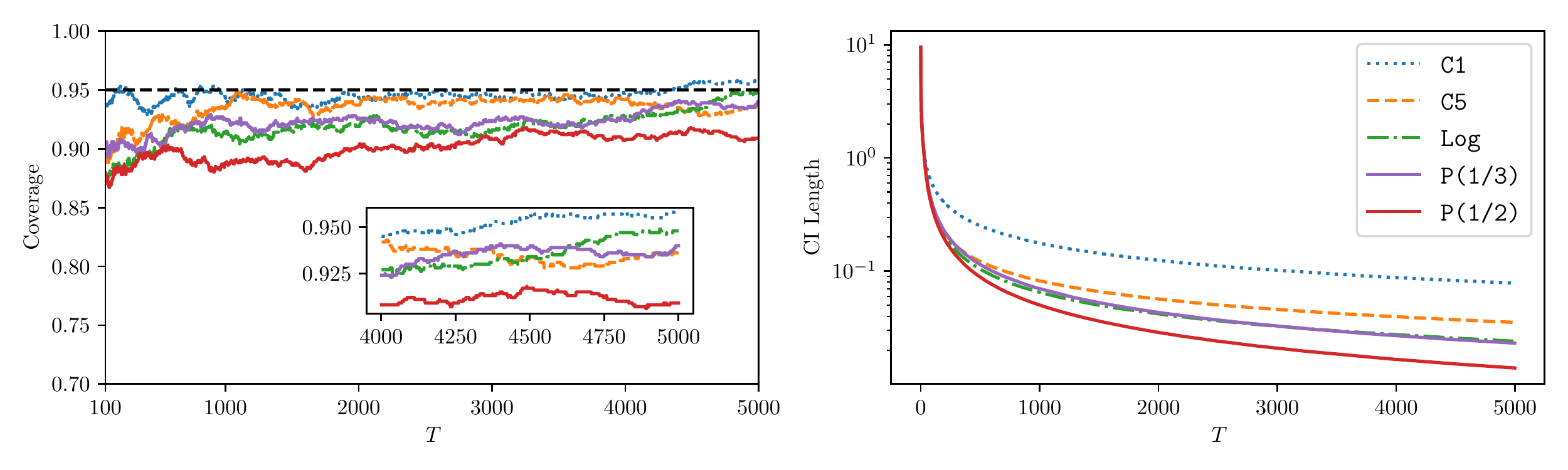}\\
	\includegraphics[width=\textwidth]{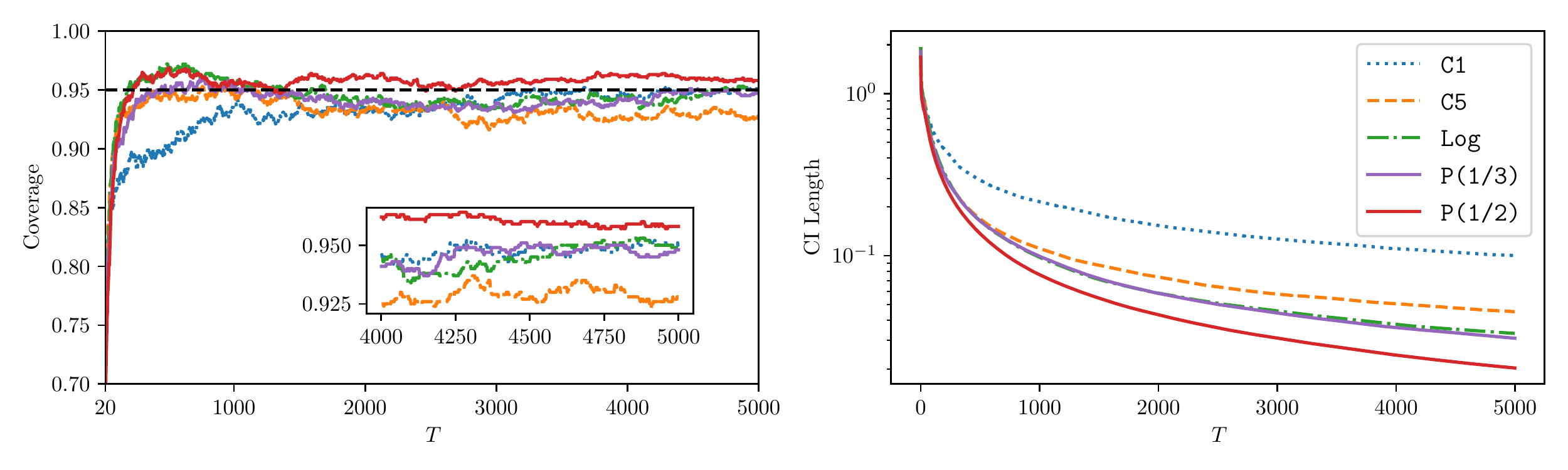}
	\caption{Comparison of the plug-in (the top row) and random scaling (the bottom row) in linear regression.
	Left: Empirical coverage rate against the number of communication. 
	Black dashed line denotes the nominal coverage rate of 95\%.
	Right: Length of confidence intervals.}
	\label{fig:linear}
\end{figure}

\begin{figure}[t!]
	\centering
	\includegraphics[width=\textwidth]{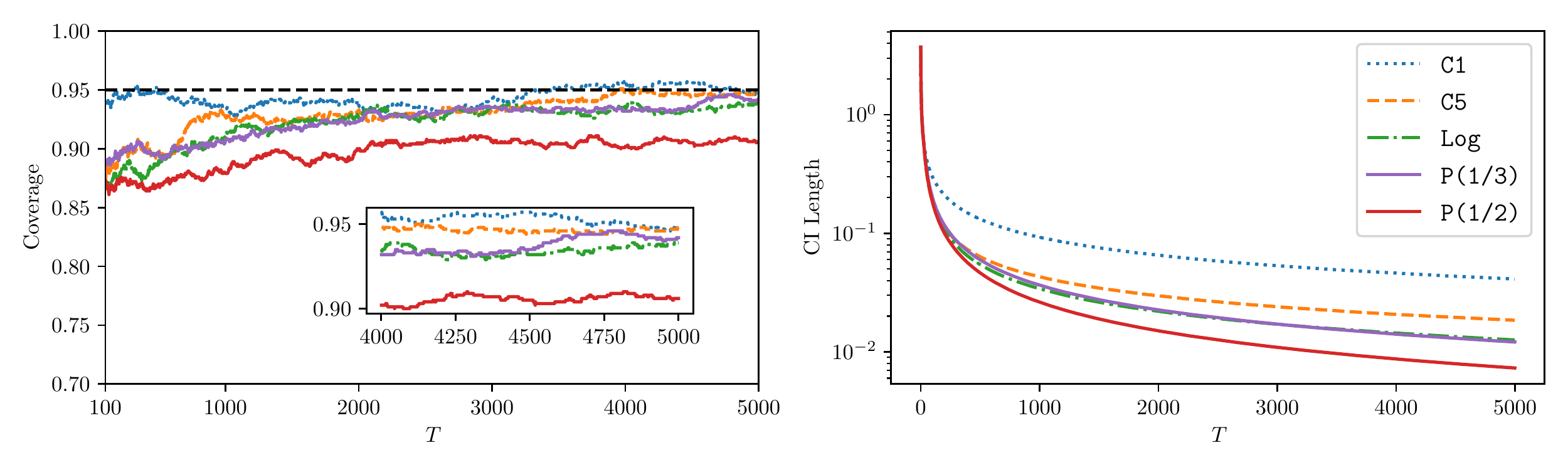}\\
	\includegraphics[width=\textwidth]{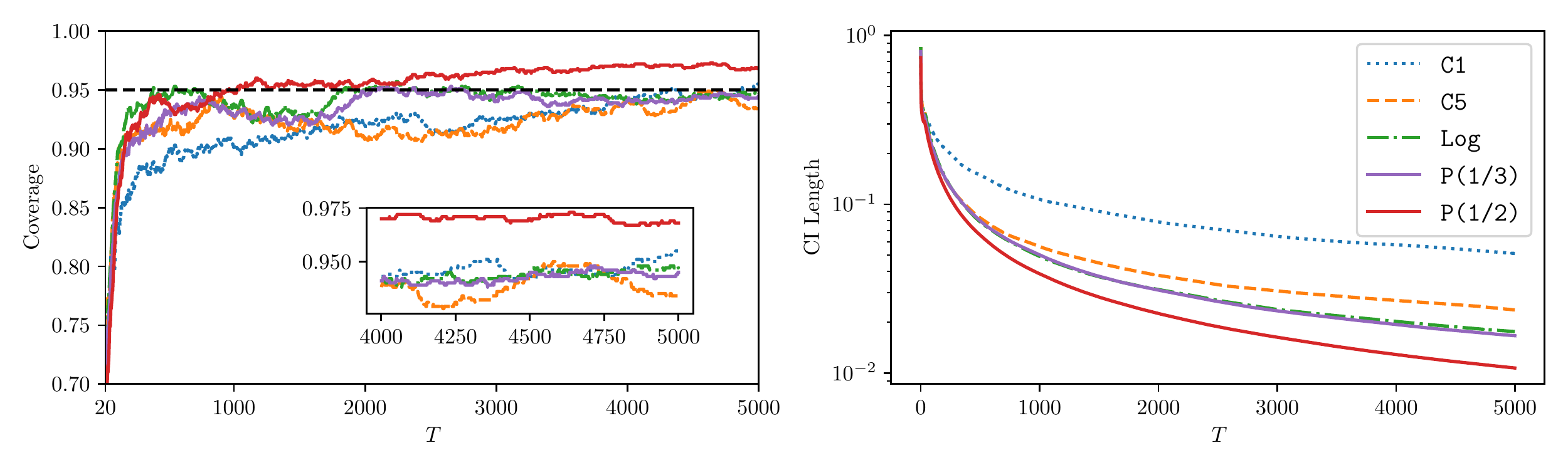}
	\caption{Comparison of the plug-in (the top row) and random scaling (the bottom row) estimators in logistic regression.
	Left: Empirical coverage rate against the number of communication. 
	Black dashed line denotes the nominal coverage rate of 95\%.
	Right: Length of confidence intervals.}
	\label{fig:logistic}
\end{figure}

\section{Numerical Simulations}
\label{sec:experiments}
This section investigates the empirical performance of the plug-in and random scaling methods via Monte Carlo experiments.
We consider both the linear and logistic regression models.
At iteration $t$, the $k$-th client observes the pair $(\a_t^k, b_t^k)$ with $\a_t^k$ the $d$-dimensional covariates generated from the multivariate normal distribution $\NM(\0, \I_d)$ and $b_t^k$ the response generated according to the model. We detail the data generation process as follows:
\begin{itemize}
\item In linear regression, $b_t^k = (\a_t^k)^\top\x_k^* + \varepsilon_t^k$ where the $\varepsilon_t^k$ are i.i.d.\ according to $\NM(\0, \I_d)$ and $\x_k^*$ is the true local parameter which we also generate from $\NM(\0, \I_d)$.
In this case, the global parameter $\x^*$ is the average of $\x_k^*$'s.
\item In logistic regression, $b_t^k  \in \{0, 1\}$ is generated to be $1$ with probability $\sigma((\a_t^k)^\top\x^*)$ and $0$ with  probability $1-\sigma((\a_t^k)^\top\x^*)$.
Here $\sigma(\theta) = 1/(1+\exp(-\theta))$ is the sigmoid function.
We do not impose data heterogeneity for logistic regression in order to avoid numerical error in the calculation of $\x^*$.
Here $\x^*$ is equi-spaced on the interval $[0, 1]$ following previous works~\citep{chen2020statistical,lee2021fast}.
\end{itemize}
We set $\gamma_m = \gamma_0/m^{0.505}$ with $\gamma_0 = 0.5$ for linear regression and $\gamma_0 = 2$ for logistic regression.
The initial value $\bx_0$ is set as zero.
We fix $K=10$ in all our experiments and vary the number of rounds $T$.
In all cases, we set $E_m =1$ for the first 5\% observations as a warm-up and then increase $E_m$ from scratch, i.e., $E_m = E_{m-5\%*T}'$ for another sequence $\{E_m'\}$.
We consider five choices of $\{E_m'\}_m$, namely $(i)$ \texttt{C1}: constant $E_m'\equiv1$, $(ii)$ \texttt{C5}: constant $E_m'\equiv 5$, $(iii)$ \texttt{Log}: logarithmic $E_m'=\lceil\log_2(m+1)\rceil$, $(iv)$ \texttt{P(1/3)}: power $E_m'=\lceil m^{1/3}\rceil$, and $(v)$ \texttt{P(1/2)}: power $E_m' = \lceil m^{1/2}\rceil$.
The nominal coverage probability is set at $95\%$.
The performance is measured by three statistics: the coverage rate, the average length of the 95\% confidence interval, and the average communication frequency. 
For brevity, we focus on the first coefficient $\x_1^*$ hereafter.
All the reported results are obtained by taking the average of 1000 independent runs.

We first turn to study the communication efficiency for Local SGD.
From Figure~\ref{fig:convergence}, we find the faster $E_m$ grows, the faster the $L_2$ convergence in terms of communication, which is consistent with previous studies from optimization perspective~\citep{mcmahan2017communication,lin2018don}.
Figure~\ref{fig:linear} shows the empirical coverage rates and confidence interval lengths in linear regression, both obtained by averaging over 1000 Local SGD paths.
The result of logistic regression is depicted in Figure~\ref{fig:logistic}.
For plug-in, though wandering above $90\%$, the faster $E_m$ family (namely, \texttt{Log}, \texttt{P(1/3)} and \texttt{P(1/2)}) has relatively inferior coverage rate than the slower $E_m$ family (namely, \texttt{C1} and \texttt{C5}).
For random scaling, it is clear that the coverage rate of all the methods fluctuates around $95\%$.
Though with a much smaller deviation from $95\%$, the slow $E_m$ family has the slower shrinkage rate for its confidence interval.
By contrast, the faster $E_m$ family achieves comparable coverage with faster shrinkage of confidence intervals.
It implies that Local SGD has high efficiency of communication and maintains a good statistic efficiency via random scaling.

 \begin{table}[t!]
	\caption{Simulation results of linear regression with $d = 5$.
	The standard errors of coverage rates $\widehat{p}$ are computed via $\sqrt{\widehat{p}(1-\widehat{p})/1000} \times 100\%$ and reported inside the parentheses.
	} 
	\centering 
	\begin{tabular}{c|l|l|cccc} 
	    \toprule
		Methods&\multicolumn{2}{c|}{Items} &  $t_T=5000$ & $t_T=10000$ & $t_T=20000$ & $t_T=40000$ \\
		\midrule
		\multirow{10}{*}{Plug-in}
		&\multirow{5}{*}{  \tabincell{c}{Cov Rate\\(\%)} }
		& \texttt{C1}& 95.70(0.641) & 94.20(0.739) & 94.20(0.739) & 93.80(0.763) \\
		&& \texttt{C5}& 93.70(0.768) & 94.00(0.751) & 94.30(0.733) & 93.10(0.801) \\
		&& \texttt{Log}& 91.70(0.872) & 93.20(0.796) & 93.80(0.763) & 93.80(0.763) \\
		&& \texttt{P(1/3)}& 91.90(0.863) & 92.70(0.823) & 93.90(0.757) & 93.60(0.774)\\
		&& \texttt{P(1/2)}& 91.10(0.900) & 92.60(0.828) & 93.90(0.757) & 93.80(0.763) \\
		\cmidrule{2-7}
		&\multirow{5}{*}{ \tabincell{c}{Avg Len\\($10^{-2}$)} }
		&\texttt{C1}& 7.857(0.099) & 5.547(0.050) & 3.917(0.025) & 2.768(0.013) \\
		&& \texttt{C5}& 9.737(0.242) & 6.868(0.121) & 4.847(0.061) & 3.423(0.031) \\
		&&\texttt{Log}& 12.168(0.371) & 8.953(0.204) & 6.602(0.106) & 4.864(0.058)\\
		&& \texttt{P(1/3)}& 11.372(0.336) & 8.656(0.195) & 6.613(0.110) & 5.059(0.063)\\
		&& \texttt{P(1/2)}& 15.431(0.559) & 12.100(0.327) & 9.433(0.188) & 7.300(0.112)\\
		\midrule
		\multirow{10}{*}{ \tabincell{c}{Random\\Scaling} }
		&\multirow{5}{*}{  \tabincell{c}{Cov Rate\\(\%)} }
		& \texttt{C1}& 95.00(0.689) & 93.90(0.757) & 93.70(0.768) & 94.80(0.702) \\
		&&\texttt{C5}& 97.70(0.474) & 96.90(0.548) & 97.20(0.522) & 96.90(0.548)\\
		&& \texttt{Log}& 98.20(0.420) & 98.70(0.358) & 98.90(0.330) & 98.80(0.344)\\
		&& \texttt{P(1/3)}& 97.60(0.484) & 98.20(0.420) & 98.50(0.384) & 98.00(0.443) \\
		&& \texttt{P(1/2)}& 96.00(0.620) & 97.20(0.522) & 96.40(0.589) & 96.60(0.573)\\
		\cmidrule{2-7}
		&\multirow{5}{*}{ \tabincell{c}{Avg Len\\($10^{-2}$)} }
		& \texttt{C1}& 10.011(4.343) & 7.081(3.106) & 5.010(2.092) & 3.605(1.511) \\
		&& \texttt{C5}& 14.434(6.950) & 10.043(4.923) & 7.078(3.389) & 4.946(2.448) \\
		&& \texttt{Log}& 19.187(9.763) & 14.120(7.154) & 10.430(5.219) & 7.611(3.895)\\
		&& \texttt{P(1/3)}& 16.781(8.397) & 12.810(6.460) & 9.821(4.906) & 7.440(3.777)\\
		&& \texttt{P(1/2)}& 20.888(10.842) & 16.127(8.004) & 12.379(6.027) & 9.314(4.460)\\
		\bottomrule
	\end{tabular}
	\label{table:linear_d5} 
\end{table}

 \begin{table}[t!]
	\caption{Simulation results of logistic regression with $d = 5$.
	The standard errors of coverage rates $\widehat{p}$ are computed via $\sqrt{\widehat{p}(1-\widehat{p})/1000} \times 100\%$ and reported inside the parentheses.
	} 
	\centering 
	\begin{tabular}{c|l|l|cccc} 
	    \toprule
		Methods&\multicolumn{2}{c|}{Items} &  $t_T=5000$ & $t_T=10000$ & $t_T=20000$ & $t_T=40000$ \\
		\midrule
		\multirow{10}{*}{Plug-in}
		&\multirow{5}{*}{  \tabincell{c}{Cov Rate\\(\%)} }
		& \texttt{C1} & 94.70(0.708) & 93.50(0.780) & 94.60(0.715) & 95.40(0.662)\\
		&& \texttt{C5} & 93.00(0.807) & 92.30(0.843) & 93.50(0.780) & 94.10(0.745)\\
		&& \texttt{Log} & 92.30(0.843) & 92.10(0.853) & 92.60(0.828) & 92.90(0.812)\\
		&& \texttt{P(1/3)} & 92.70(0.823) & 92.00(0.858) & 92.50(0.833) & 92.90(0.812)\\
		&& \texttt{P(1/2)} & 90.80(0.914) & 92.20(0.848) & 91.70(0.872) & 92.10(0.853)\\
		\cmidrule{2-7}
		&\multirow{5}{*}{ \tabincell{c}{Avg Len\\($10^{-2}$)} }
		& \texttt{C1} & 4.113(0.046) & 2.903(0.022) & 2.049(0.011) & 1.448(0.005)\\
		&& \texttt{C5} & 5.081(0.118) & 3.587(0.057) & 2.534(0.029) & 1.790(0.014)\\
		&& \texttt{Log} & 6.347(0.175) & 4.681(0.093)& 3.453(0.049) & 2.544(0.027)\\
		&& \texttt{P(1/3)} & 5.949(0.146) &  4.526(0.091)& 3.456(0.049)  & 2.647(0.027)\\
		&& \texttt{P(1/2)} & 8.062(0.256) &  6.320(0.149)& 4.927(0.088)  & 3.821(0.052)\\
		\midrule
		\multirow{10}{*}{ \tabincell{c}{Random\\Scaling} }
		&\multirow{5}{*}{  \tabincell{c}{Cov Rate\\(\%)} }
		& \texttt{C1} & 95.50(0.656) & 92.40(0.838) & 94.10(0.745) & 94.70(0.708)\\
		&& \texttt{C5} & 96.00(0.620) & 95.90(0.627) & 96.80(0.557) & 95.80(0.634)\\
		&& \texttt{Log} & 97.60(0.484) & 97.40(0.503) & 97.80(0.464) & 98.20(0.420)\\
		&& \texttt{P(1/3)} & 96.10(0.612) & 96.60(0.573) & 97.50(0.494) & 97.90(0.453)\\
		&& \texttt{P(1/2)} & 94.40(0.727) & 94.30(0.733) & 94.50(0.721) & 95.10(0.683)\\
		\cmidrule{2-7}
		&\multirow{5}{*}{ \tabincell{c}{Avg Len\\($10^{-2}$)} }
		& \texttt{C1} & 5.112(2.302) & 3.612(1.502) & 2.646(1.162) & 1.877(0.816)\\
		&& \texttt{C5} & 7.296(3.714) & 5.166(2.535) & 3.687(1.836) & 2.637(1.316)\\
		&& \texttt{Log} & 9.703(5.176) & 7.241(3.713) & 5.383(2.787) & 4.023(2.063)\\
		&& \texttt{P(1/3)} & 8.499(4.465) & 6.569(3.345) & 5.071(2.621) & 3.924(1.999)\\
		&& \texttt{P(1/2)} & 10.574(5.688) & 8.278(4.193) & 6.340(3.194) & 4.880(2.366)\\
		\bottomrule
	\end{tabular}
	\label{table:logistic_d5} 
\end{table}

We then turn to the empirical performance of Local SGD with limited computation or finite samples.
Table~\ref{table:linear_d5} shows the empirical performance of the five methods under linear models with four different $t_T$'s.
$t_T$ is actually the total iteration each client runs through $T$ rounds or equivalently the number of observations they receive.
From the table, all the methods achieve good performance.
The random scaling gives better average coverage rates than the plug-in method, because its average coverage rates of all different communication intervals are near (or even exceed) $95\%$.
However, its average length is usually larger than that of plug-in.
Furthermore, its average length usually has a much larger deviation than that of plug-in.
For example, when $t_T=5000$, for $\texttt{C5}$, the standard deviation of average lengths for plug-in is $0.807 \times 10^{-2}$, while it increases to $3.714 \times 10^{-2}$ for random scaling.
Such a wider average length might account for the unexpected advantage on the average coverage rates.
However, as the communication round increases and more observations are available, both the average length and its deviations decrease.

In addition, comparing the results of \texttt{Log}, \texttt{P(1/3)}, and \texttt{P(1/2)}, we can find that the faster $E_m$ increases, the larger average length as well as its standard deviations. However, they all have satisfactory performance when observations are sufficient.
Indeed, Local SGD trades more computation for less communication, resulting in a residual error gradually accumulated when communication is off, slowing down the convergence rate and enlarging asymptotic variance (e.g., the existence of $\nu$).
However, the benefit is also attractive: the averaged communication frequency is substantially reduced and the convergence in terms of communication largely increases.
It implies that Local SGD obtains both statistical efficiency and communication efficiency as expected.
We further consider the logistic regression, which is a standard non-linear model.
The result is given in Table~\ref{table:logistic_d5}.
A similar pattern is observed: random scaling has higher average coverage rates at the price of wider average lengths which typically shrink as more observations are generated.

\section{Conclusion and Future Work}
\label{sec:conclusion}

This paper studies how to perform statistical inference via Local SGD in federated learning.
We have established a functional central limit theorem for the averaged iterates of Local SGD and presented two fully online inference methods.
We have shown that the Local SGD has statistical efficiency with its asymptotic variance achieving the Cram\'{e}r–Rao lower bound and communication efficiency with the averaged communication efficiency vanishing asymptotically.

There are many interesting issues for future work.
One is to relax the current assumptions and consider Local SGD for more challenging optimization problems (e.g., non-smooth or non-convex problems).
Our theory shows that Local SGD enjoys statistical optimality in an asymptotic sense, and it is definitely not also optimal in finite-time convergence~\citep{woodworth2021min}. 
It is then interesting to analyze the statistical properties of other state-of-the-art algorithms in federated learning.
For example, \citet{karimireddy2019scaffold} proposed a new algorithm using control variates to remove the effect of data heterogeneity, which achieves a better non-asymptotic convergence rate.
It is also interesting to devise more powerful algorithms as well inference methods to handle the challenge in the decentralized big data era~\citep{fan2021modern}.

\bibliography{bib/optimization,bib/federated,bib/distributed,bib/stat}
\bibliographystyle{plainnat}
\appendix
\begin{appendix}
	\onecolumn
	\begin{center}
		{\huge {Supplementary Material to "Statistical Estimation and Inference via Local SGD in Federated Learning"}}
	\end{center}

\section{Proofs for the FCLT}
\label{append:main-proof}

This appendix provides a self-contained proof of Theorem~\ref{thm:fclt} as well as the first statement of Theorem~\ref{thm:clt}.

\subsection{Main Proof}

We follows the perturbed iterate framework that is derived by~\citet{mania2017perturbed} and widely used in recent works~\citep{stich2018local,stich2018sparsified,li2019convergence,bayoumi2020tighter,koloskova2020unified,woodworth2020local,woodworth2020minibatch}.
Then we define a virtual sequence $\bx_t$ in the following way:
\[
\bx_t = \sum_{k=1}^K p_k \x_{t}^k.
\]

Fix a $m \ge 0$ and consider $t_m \le t < t_{m+1}$. 
Local SGD yields that for any device $k \in [K]$,
\begin{eqnarray*}
\x_{t+1}^k &=& \x_t^k - \eta_m \nabla f_k(\x_t^k; \xi_t^k), \\
\x_{t_{m+1}}^k &=& \sum_{k=1}^K p_k \left( \x_{t_{m+1}-1}^k- \eta_m  \nabla f_k(\x_{t_{m+1}-1}^k; \xi_{t_{m+1}-1}^k) \right),
\end{eqnarray*}
which implies that we always have
\begin{equation}
\label{eq:average_x}
\bx_{t+1} = \bx_{t} -  \eta_m \bg_t, 
\quad \text{where} \quad \bg_t  = \sum_{k=1}^K p_k \nabla f_k(\x_t^k; \xi_t^k).
\end{equation}
Define $\s_m = \bx_{t_m} - \x^*$ and recall that $E_m = t_{m+1}-t_m$ and $\gamma_m = \eta_m E_m$.
Iterating~\eqref{eq:average_x} from $t=t_m$ to $t_{m+1}-1$ gives
\begin{equation}
\label{eq:iterate_s}
\s_{m+1} = \s_m - \eta_m \sum_{t=t_m}^{t_{m+1}-1} \bg_t = \s_m  - \gamma_m \sv_m,
\quad \text{where} \quad 
\sv_m = \frac{1}{E_m}  \sum_{t=t_m}^{t_{m+1}-1} \bg_t.
\end{equation}
We further decompose $\sv_m$ into four terms.
\begin{align}
\sv_m &= \sG\s_m +
\left(\nabla f (\bx_{t_m}) -\sG\s_m \right) + (\sh_m - \nabla f (\bx_{t_m})) + (\sv_m - \sh_m) \nonumber \\
&:=\sG\s_m + \sr_m  + \seps_m + \sdelta_m \label{eq:decompose_v}
\end{align}
where $\sG = \nabla^2 f(\x^*)$ is the Hessian at the optimum $\x^*$ which is non-singular from our assumption, and 
\begin{equation}
\label{eq:}
\sh_m = \frac{1}{E_m}  \sum_{t=t_m}^{t_{m+1}-1} \sum_{k=1}^K p_k \nabla f_k(\bx_{t_m}; \xi_t^k).
\end{equation}
Note that $\sh_m$ is almost identical to $\sv_m$ except that all the stochastic gradients in $\sh_m$ are evaluated at $\bx_{t_m}$ while those in $\sv_m$ are evaluated at local variables $\x_t^k$'s.

Making use of~\eqref{eq:iterate_s} and~\eqref{eq:decompose_v}, we have 
\begin{equation}
\label{eq:main_iterate}
\s_{m+1} = (\sI - \gamma_m\sG) \s_m - \gamma_m(\sr_m + \seps_m + \sdelta_m) := \sB_m \s_m - \gamma_m\sU_m,
\end{equation}
where $\sB_m := \sI - \gamma_m\sG$ and $\sU_m := \sr_m + \seps_m + \sdelta_m$ for short.
Recurring~\eqref{eq:main_iterate} gives
\begin{align}
\label{eq:main_iterate_explicit}
\s_{m+1}=\left(\prod_{j=0}^{m}\sB_j\right) \s_0 - \sum\limits_{j=0}^{m}\left(\prod\limits_{i=j+1}^{m}\sB_i\right) \gamma_j\U_j.
\end{align}
Here we use the convention that $\prod\limits_{i=m+1}^{m}\sB_i = \sI$ for any $m \ge 0$.

For any $r \in [0, 1]$ and $T \ge 1$, define 
\begin{equation}
\label{eq:h_r_T}
h(r, T) = \max\left\{ n \in \ZB_+ \bigg|  r \sum_{m=1}^T \frac{1}{E_m}  \ge  \sum_{m=1}^n \frac{1}{E_m}  \right\}.
\end{equation}
From Assumption~\ref{asmp:E}, we know that $\sum_{m=1}^T \frac{1}{E_m}  \to \infty$ as $T \to \infty$, which implies $h(r, T) \to \infty$ meanwhile. 
Summing~\eqref{eq:main_iterate_explicit} from $m=0$ to $h(r, T)$ gives
\begin{align}
\label{eq:main_iterate_1}
\frac{\sqrt{ t_{T}}}{T}\sum_{m=0}^{h(r, T) } \s_{m+1}
&=\frac{\sqrt{t_{T}}}{T}\sum_{m=0}^{h(r, T) }\left[\left(\prod_{j=0}^{m}\sB_j\right) \s_0 - \sum\limits_{j=0}^{m}\left(\prod\limits_{i=j+1}^{m}\sB_i\right) \gamma_j\U_j\right] \nonumber \\
&=\frac{\sqrt{t_{T}}}{T}\sum_{m=0}^{h(r, T)}\left(\prod_{j=0}^{m}\sB_j\right) \s_0  -\frac{\sqrt{t_{T}}}{T}
\sum_{j=0}^{h(r, T) }\sum\limits_{m=j}^{h(r, T)}\left(\prod\limits_{i=j+1}^{m}\sB_i\right) \gamma_j\U_j.
\end{align}

\begin{lem}[Lemma 1 in~\citet{polyak1992acceleration}]
	\label{lem:A}
	Recall that $\sB_i := \sI - \gamma_i\sG$ and $\sG$ is non-singular.
	For any $n \ge j$, define $\sA_{j}^{n}$ as
	\begin{equation}
	\label{eq:A}
		\sA_{j}^n = \sum\limits_{l=j}^{n}\left(\prod\limits_{i=j+1}^{l}\sB_i\right) 
	\gamma_j.
	\end{equation}
	Under Assumption~\ref{asmp:gamma}, there exists some universal constant $C_0 > 0$ such that for any $n\ge j\ge 0$, $\|\sA_{j}^{n}\| \le C_0$.
	Furthermore, it follows that $\lim\limits_{n \to \infty} \frac{1}{n} \sum_{j=0}^{n}\|\sA_{j}^{n} - \sG^{-1}\| = 0.$ 
\end{lem}
Using the notation of $\sA_{j}^n$, we can further simplify~\eqref{eq:main_iterate_1} as
\[
\frac{\sqrt{t_T}}{T}\sum_{m=0}^{h(r, T) } \s_{m+1}
=\frac{\sqrt{t_T}}{T \gamma_0} \sA_0^{h(r, T)} \sB_0\s_0
- \frac{\sqrt{t_T}}{T}
\sum_{m=0}^{h(r, T) } \sA_m^{h(r, T)} \sU_m.
\]
Since $\sU_m = \sr_m + \seps_m + \sdelta_m$, then
\begin{align*}
\frac{\sqrt{t_T}}{T}\sum_{m=0}^{h(r, T) } \s_{m+1} 
+ \frac{\sqrt{t_{T}}}{T} \sum_{m=0}^{h(r, T)}\sG^{-1} \seps_m 
&=\frac{\sqrt{t_T}}{T \gamma_0} \sA_0^{h(r, T)} \sB_0\s_0
- \frac{\sqrt{t_T}}{T}
\sum_{m=0}^{h(r, T) } \sA_m^{h(r, T)} (\sr_m +  \sdelta_m)\\
& \qquad -\frac{\sqrt{t_T}}{T}
\sum_{m=0}^{h(r, T) } (\sA_m^{T} -\sG^{-1}) \seps_m\\
& \qquad -\frac{\sqrt{t_T}}{T}
\sum_{m=0}^{h(r, T) } (\sA_m^{h(r, T)} -\sA_m^{T} ) \seps_m\\
&:= \TM_0  - \TM_1 - \TM_2 - \TM_3,
\end{align*}
where for simplicity we denote
\begin{gather*}
\TM_0 = \frac{\sqrt{t_T}}{T \gamma_0} \sA_0^{h(r, T)} \sB_0\s_0, \quad
\TM_1 = \frac{\sqrt{t_T}}{T}
\sum_{m=0}^{h(r, T) } \sA_m^{h(r, T)} (\sr_m +  \sdelta_m), \\
\TM_2 = \frac{\sqrt{t_T}}{T}
\sum_{m=0}^{h(r, T) } (\sA_m^{T} -\sG^{-1}) \seps_m, \quad
\TM_3 = \frac{\sqrt{t_T}}{T}
\sum_{m=0}^{h(r, T) } (\sA_m^{h(r, T)} -\sA_m^{T} ) \seps_m.
\end{gather*}
With the last equation, we are ready to prove the main theorem which illustrates the partial-sum asymptotic behavior of $\frac{\sqrt{t_T}}{T}\sum_{m=0}^{h(r, T) } \s_{m+1} $.
The main idea is that we first figure out the partial-sum asymptotic behavior of $\frac{\sqrt{t_{T}}}{T} \sum_{m=0}^{h(r, T)}\sG^{-1} \seps_m$ and then show that their difference is uniformly small, i.e.,
\[
\sup_{r \in [0 ,1]}\left\|\frac{\sqrt{t_T}}{T}\sum_{m=0}^{h(r, T) } \s_{m+1} 
+ \frac{\sqrt{t_{T}}}{T} \sum_{m=0}^{h(r, T)}\sG^{-1} \seps_m \right\| = o_{\PB}(1).
\]
For the second step, it suffices to show that the four separate terms: $\sup_{r \in [0, 1]} \|\TM_0\|, \sup_{r \in [0, 1]} \|\TM_1\|$, $\sup_{r \in [0, 1]} \|\TM_2\|$, and $\sup_{r \in [0, 1]} \|\TM_4\|$ are  $o_{\PB}(1)$, respectively.
With this idea, our following proof is naturally divided into fives parts.

The establishment of almost sure and $L_2$ convergence in Lemma~\ref{lem:a.s.convergence} will ease our proof. 
The following lemma proves the first statement of Theorem~\ref{thm:clt}.
The second statement of Theorem~\ref{thm:clt} follows directly from Theorem~\ref{thm:fclt} which we are going to prove via an argument of the continuous mapping theorem.

\begin{lem}[Almost sure and $L_2$ convergence]
	\label{lem:a.s.convergence}
	Under Assumptions~\ref{asmp:convex},~\ref{asmp:noise}, and~\ref{asmp:gamma}, $\bx_{t_m} \to \x^*$ almost surely when $m$ goes to infinity.
	In addition, there exists some $\widetilde{C}_0 > 0$ such that
	\[
	\EB\|\bx_{t_m} - \x^*\|^2 \le \widetilde{C}_0 \gamma_m.
	\]
\end{lem}

\paragraph{Part 1: Partial-sum asymptotic behavior of $\frac{\sqrt{t_{T}}}{T} \sum_{m=0}^{h(r, T)}\sG^{-1} \seps_m$.}

\begin{lem}
	\label{lem:T1}
	Under Assumptions~\ref{asmp:convex},~\ref{asmp:noise},~\ref{asmp:gamma} and~\ref{asmp:E}, the functional martingale CLT holds, namely, for any $r \in [0, 1]$,
	\[
	\frac{\sqrt{t_{T}}}{T} \sum_{m=0}^{h(r, T)}  \sG^{-1}\seps_m \Rightarrow  \sqrt{\nu}  \sG^{-1}\sS^{1/2} \B_d(r), 
	\]
	where $h(r, T)$ is defined in~\eqref{eq:h_r_T}
	and $\B_d(r)$ is the $d$-dimensional standard Brownian motion. 
\end{lem}

\paragraph{Part 2: Uniform negligibility of $\TM_0$.}
Lemma~\ref{lem:A} characterizes the asymptotic behavior of $\sA_{j}^n$.
It is uniformly bounded.
It implies 
\[
\sup_{r \in [0, 1]} \|\TM_0\| = 
\frac{\sqrt{t_T}}{T \gamma_0} \sup_{r \in [0, 1]} \|\sA_0^{h(r, T)} \sB_0\s_0\| \le  \frac{\sqrt{t_T}}{T \gamma_0} C_0\|\sB_0\s_0\| \to 0,
\]
as a result of $\frac{\sqrt{t_T}}{T} \to 0$ when $T \to \infty$.
\paragraph{Part 3: Uniform negligibility of $\TM_1$.}
The uniform boundedness of $\sA_{j}^n$ implies
\begin{align*}
\sup_{r \in [0, 1]} \|\TM_1\| 
&=\sup_{r \in [0, 1]}\frac{\sqrt{t_T}}{T}
\left\|\sum_{m=0}^{h(r, T) } \sA_m^{h(r, T)} (\sr_m +  \sdelta_m)\right\|\\
&\le \sup_{r \in [0, 1]}\frac{\sqrt{t_T}}{T}\sum_{m=0}^{h(r, T) } C_0 (\|\sr_m\| +  \|\sdelta_m\|)\\
&=\frac{\sqrt{t_T}}{T}\sum_{m=0}^{T} C_0 (\|\sr_m\| +  \|\sdelta_m\|),
\end{align*}
where the last inequality uses the fact that $h(r, T)$ increases in $r$ and $h(1, T) = T$.
The following two lemmas together imply that $\sup_{r \in [0, 1]} \|\TM_1\|  = o_{\PB}(1)$.

\begin{lem}
	\label{lem:T2}
	Under Assumptions~\ref{asmp:convex},~\ref{asmp:noise} and~\ref{asmp:gamma},  we have that
	\[
	\frac{\sqrt{t_{T}}}{T}\sum_{m=0}^T \left\|  \sr_m\right\| = o_{\PB}(1).
	\]
\end{lem}

\begin{lem}
	\label{lem:T3}
	Under Assumptions~\ref{asmp:convex},~\ref{asmp:noise} and~\ref{asmp:gamma}, we have that
	\[
	\frac{\sqrt{t_{T}}}{T} \sum_{m=0}^T \left\|\sdelta_m\right\| = o_{\PB}(1).
	\]
\end{lem}

\paragraph{Part 4: Uniform negligibility of $\TM_2$.}
By Doob's maximum inequality, it follows that
\begin{align*}
\EB\sup_{r \in [0, 1]} \|\TM_2 \|^2
&=\EB\sup_{r \in [0, 1]}   \frac{t_T}{T^2} \left\|
\sum_{m=0}^{h(r, T) } (\sA_m^{T} -\sG^{-1}) \seps_m\right\|^2\\
&\le \frac{t_T}{T^2} \EB\left\|
\sum_{m=0}^{T} (\sA_m^{T} -\sG^{-1}) \seps_m\right\|^2\\
&=\frac{t_T}{T^2} \sum_{m=0}^{T} \EB \left\|
(\sA_m^{T} -\sG^{-1}) \seps_m\right\|^2\\
&\le\frac{t_T}{T^2} \sum_{m=0}^{T}\left\|
\sA_m^{T} -\sG^{-1}\right\|^2 \EB\left\|  \seps_m\right\|^2.
\end{align*}
Because $\seps_m = \sh_m - \nabla f(\bx_{t_{m}}) = \frac{1}{E_m}  \sum_{t=t_m}^{t_{m+1}-1} \left(\nabla f(\bx_{t_m}; \xi_t) - \nabla f(\bx_{t_m})\right)$ is the mean of $E_m$ i.i.d.\ copies of $\eps(\bx_{t_{m}}) := \nabla f(\bx_{t_{m}}; \xi_{t_m}) - \nabla f(\bx_{t_{m}})$ at a fixed $\bx_{t_{m}}$, it implies that
\begin{equation}
\label{eq:bound_noise}
\EB\left\|  \seps_m\right\|^2 
= \frac{1}{E_m} \EB\|\eps(\bx_{t_{m}})\|^2 \le  \frac{1}{E_m} \left( C_1 + C_2 \EB\|\bx_{t_{m}} - \x^*\|^2 \right) \precsim \frac{1}{E_m},
\end{equation}
where the first inequality is from Lemma~\ref{lem:bound_noise} with $C_1, C_2$ two universal constants defined therein and the second inequality uses Lemma~\ref{lem:a.s.convergence}.
Using the last result, we have that
\[
\EB\TM_{2} \precsim \frac{t_T}{T^2} \sum_{m=0}^{T} \frac{1}{E_m}\left\|
\sA_m^{T} -\sG^{-1}\right\|^2.
\]
By Lemma~\ref{lem:A}, it follows that as $T \to \infty$,
\[
\frac{1}{T}\sum_{m=0}^{T}\left\|\sA_m^{T} -\sG^{-1}\right\|^2 
\le (C_0 + \|\sG^{-1}\|) \cdot \frac{1}{T}\sum_{m=0}^{T}\left\|\sA_m^{T} -\sG^{-1}\right\| \to 0.
\]
Lemma~\ref{lem:iterate} implies that $\EB\sup_{r \in [0, 1]} \|\TM_2 \|^2 = o(1)$.

\begin{lem}
	\label{lem:iterate}
	Let $\{E_m\}$ be the positive-integer-valued sequence that satisfies Assumption~\ref{asmp:E}.
	Let $\{a_{m, T}\}_{m\in [T], T\ge 1}$ be a non-negative uniformly bounded sequence satisfying $\lim\limits_{T \to \infty} \frac{1}{T}\sum_{m=0}^{T-1} a_{m, T} = 0$.
	Then
	\[
	\lim\limits_{T \to \infty} \frac{(\sum_{m=0}^{T-1} E_m)(\sum_{m=0}^{T-1} E_m^{-1}a_{m, T})}{T^2}  = 0.
	\]
\end{lem}

\paragraph{Part 5: Uniform negligibility of $\TM_3$.}
It is subtle to handle $\TM_3$ because its coefficient depends on $r$.
\begin{align*}
\|\TM_3\| 
&= \frac{\sqrt{t_T}}{T}\left\| \sum_{m=0}^{h(r, T) } (\sA_m^{T}-\sA_m^{h(r, T)}) \seps_m\right\|\\
&=\frac{\sqrt{t_T}}{T}\left\| \sum_{m=0}^{h(r, T) } \sum_{l=h(r, T)+1}^T \left(\prod\limits_{i=m+1}^{l}\sB_{ i}\right) \gamma_m  \seps_m\right\|\\
&=\frac{\sqrt{t_T}}{T}\left\| \sum_{l=h(r, T)+1}^T
\sum_{m=0}^{h(r, T) }  \left(\prod\limits_{i=m+1}^{l}\sB_{i}\right) \gamma_m  \seps_m\right\|\\
&=\frac{\sqrt{t_T}}{T}\left\| \sum_{l=h(r, T)+1}^T  \left(\prod\limits_{i=h(r, T)+1}^{l}\sB_i\right) \sum_{m=0}^{h(r, T) } \left(\prod\limits_{i=m+1}^{h(r, T)}\sB_i\right) \gamma_m  \seps_m\right\|\\
&\precsim \frac{\sqrt{t_T}}{T}\left\|\frac{1}{\gamma_{h(r, T)+1}}\sum_{m=0}^{h(r, T) } \left(\prod\limits_{i=m+1}^{h(r, T)}\sB_i\right) \gamma_m  \seps_m\right\|,
\end{align*}
where the last inequality uses
\[
\left\| \sum_{l=h(r, T)+1}^T  \left(\prod\limits_{i=h(r, T)+1}^{l}\sB_i\right)\gamma_{h(r, T)+1} \right\| = \left\| \sA_{h(r, T)+1}^T \sB_{h(r, T)+1} \right\| \precsim 1.
\]
Lemma~\ref{lem:sup_epsi} shows that $\sup_{r \in [0, 1]}\|\TM_3\| = o_{\PB}(1)$.
\begin{lem}
	\label{lem:sup_epsi}
Under Assumptions~\ref{asmp:noise} and~\ref{asmp:E}, it follows that
\[
\sup_{r \in [0, 1]}\frac{\sqrt{t_T}}{T}\left\|\frac{1}{\gamma_{h(r, T)+1}}\sum_{m=0}^{h(r, T) }  \left(\prod\limits_{i=m+1}^{h(r, T)}\sB_i\right) \gamma_m  \seps_m\right\| = o_{\PB}(1).
\]
\end{lem}

\begin{rem}
There is a more user-friendly version of Lemma~\ref{lem:sup_epsi} for a plug-and-play use.
Define an auxiliary sequence $\{ \Y_m \}_{m\ge0}$ as following: $\Y_0 = \0$ and for $m\ge 0$,
\begin{equation}
\label{eq:Y}
\Y_{m+1} = \sB_m \Y_m + \gamma_m \seps_m 
= (\sI - \gamma_m\sG) \Y_m + \gamma_m\seps_m.
\end{equation}
It is easy to verify that
\[
\Y_{t+1} = \sum_{t=0}^t \left(\prod\limits_{i=m+1}^{t}\sB_i\right) \gamma_m  \seps_m.
\]
Under this notation, Lemma~\ref{lem:sup_epsi} is equivalent to 
\[
\sup_{0 \le t \le T} \frac{\sqrt{t_T}}{T} \frac{\|\Y_{t+1}\|}{\gamma_{t+1}} = o_{\PB}(1).
\]
More formally, we have the following lemma which one can prove from Lemma~\ref{lem:sup_epsi}.
\begin{lem}
	If the martingale difference sequence $\{ \seps_m  \}_{m \ge 0}$ satisfies $ \sup_{m \ge 0}\EB\|\seps_m\|^{2 +\delta} < \infty$ for some $\delta > 0$ and Assumption~\ref{asmp:E} holds with $E_m \equiv 1$, for the sequence $\{ \Y_m \}_{m\ge0}$ defined in~\eqref{eq:Y}, we have
	\[
	\sup_{0 \le t \le T} \frac{1}{\sqrt{T}} \frac{\|\Y_{t+1}\|}{\gamma_{t+1}} = o_{\PB}(1).
	\]
\end{lem}
\end{rem}

\subsection{Proof of Lemma~\ref{lem:a.s.convergence}}
Define $\FM_t = \sigma(\{ \xi_{\tau}^k \}_{1 \le k  \le K,0\le  \tau < t})$ by the natural filtration generated by $\xi_{\tau}^k$'s, so $\{\x_t^k\}_t$ is adapted to $\{\FM_t\}_t$ and $\{\bx_{t_{m}}\}_m$ is adapted to $\{\FM_{t_m}\}_{m}$.
Notice that $\sv_m = \sh_m  + \sdelta_m$ where 
\[
\sh_m = \frac{1}{E_m}  \sum_{t=t_m}^{t_{m+1}-1} \nabla f(\bx_{t_{m}}; \xi_t)
\quad \text{and} \quad
\nabla f(\bx_{t_m}; \xi_t) = \sum_{k=1}^Kp_k\nabla f(\bx_{t_{m}}; \xi_t^k),
\]
implying $\EB[\sh_m|\FM_{t_m}] = \nabla f(\bx_{t_{m}})$.
The $L$-smoothness of $f(\cdot)$ gives that
\begin{align*}
f(\bx_{t_{m+1}})
&\le f(\bx_{t_{m}}) + \langle \nabla f(\bx_{t_{m}}),\bx_{t_{m+1}}-\bx_{t_{m}}\rangle + \frac{L}{2}\|\bx_{t_{m+1}}-\bx_{t_{m}}\|^2 \\
&= f(\bx_{t_{m}}) - \gamma_m\langle \nabla f(\bx_{t_{m}}),\sv_m\rangle + \frac{\gamma_m^2L}{2}\|\sv_m\|^2.
\end{align*}
Conditioning on $\FM_{t_m}$ in the last inequality gives
\begin{align}
\label{eq:iterate_fx}
\EB&[f(\bx_{t_{m+1}})|\FM_{t_m}]\nonumber \\
&\le f(\bx_{t_{m}}) - \gamma_m\langle \nabla f(\bx_{t_{m}}),\EB[\sv_m|\FM_{t_m}]\rangle + \frac{\gamma_m^2L}{2}\EB[\|\sv_m\|^2|\FM_{t_m}]\nonumber \\
&= f(\bx_{t_{m}}) - \gamma_m \|\nabla f(\bx_{t_{m}})\|^2  - \gamma_m\langle \nabla f(\bx_{t_{m}}),\EB[\sdelta_m|\FM_{t_m}]\rangle + \frac{\gamma_m^2L}{2}\EB[\|\sh_m + \sdelta_m\|^2|\FM_{t_m}] \nonumber\\
&\le f(\bx_{t_{m}}) - \gamma_m \|\nabla f(\bx_{t_{m}})\|^2 
+ \frac{\gamma_m}{2}\|\nabla f(\bx_{t_{m}})\|^2
+ \frac{\gamma_m}{2}\|\EB[\sdelta_m|\FM_{t_m}]\|^2\nonumber\\
& \qquad \qquad + \gamma_m^2L\EB[\|\sh_m \|^2|\FM_{t_m}]
+ \gamma_m^2L\EB[\|\sdelta_m\|^2|\FM_{t_m}]\nonumber\\
&= f(\bx_{t_{m}}) - \frac{\gamma_m}{2} \|\nabla f(\bx_{t_{m}})\|^2 + \gamma_m^2L\EB[\|\sh_m \|^2|\FM_{t_m}]  + \left(  \frac{\gamma_m}{2} + \gamma_m^2L\right) \EB[\|\sdelta_m\|^2|\FM_{t_m}],
\end{align}
where we use the conditional Jensen's inequality $\|\EB[\sdelta_m|\FM_{t_m}]\|^2  \le \EB[\|\sdelta_m\|^2|\FM_{t_m}]$.

We then bound the last two terms in the right hand side of~\eqref{eq:iterate_fx}.
\paragraph{Part 1:}
For $\EB[\|\sh_m \|^2|\FM_{t_m}]$, it follows that
\begin{align*}
\EB[\|\sh_m \|^2|\FM_{t_m}]
&=\|\EB[\sh_m|\FM_{t_m}]\|^2 + \EB[\|\sh_m - \EB[\sh_m|\FM_{t_m}] \|^2|\FM_{t_m}]   \\
&= \|\nabla f(\bx_{t_m})\|^2 +
\EB[\|\sh_m - \nabla f(\bx_{t_m}) \|^2|\FM_{t_m}]\\
&=\|\nabla f(\bx_{t_m})\|^2 +
\frac{1}{E_m} \EB[\|\nabla f(\bx_{t_m}; \xi_{t_m}) - \nabla f(\bx_{t_m}) \|^2|\FM_{t_m}],
\end{align*}
where the last equality uses the fact that $\sh_m$ is the mean of $E_m$ i.i.d.\ copies of $\nabla f(\bx_{t_m}; \xi_{t_m})  := \sum_{k=1}^K p_k \nabla f_k(\bx_{t_m}; \xi_{t_m}^k)$ given $\FM_{t_m}$, so its conditional variance is $E_m$ times smaller than the latter, 
\begin{equation}
\label{eq:variance_h}
\EB[\|\sh_m - \nabla f(\bx_{t_m}) \|^2|\FM_{t_m}] = 
\frac{1}{E_m} \EB[\|\nabla f(\bx_{t_m}; \xi_{t_m}) - \nabla f(\bx_{t_m}) \|^2|\FM_{t_m}].
\end{equation}

\begin{lem}
	\label{lem:bound_noise}
	Recall that $\eps(\bx_{t_m}) := \nabla f(\bx_{t_m}; \xi_{t_m})  - \nabla f(\bx_{t_m})$ and $\eps_k(\x_{t}^k) := \nabla f(\x_t^k; \xi_t^k)  - \nabla f(\x_t^k)$.
	Under Assumption~\ref{asmp:noise},	it follows that
	\[
	\EB_{\xi_{t}^k} \|\eps_k(\x_{t}^k)\|^2 \le C_1 + C_2 \|\x_{t}^k-\x^*\|^2
	\quad \text{and} \quad
	\EB_{\xi_{t_m}} \|\eps(\bx_{t_m})\|^2 \le C_1 + C_2 \|\bx_{t_m}-\x^*\|^2
	\]
	where $C_1 = d\max_{k \in [K]}\|\sS_k\| + \frac{dC}{2}$ and $C_2 = \frac{3dC}{2}$ with $C$ defined in Assumption~\ref{asmp:noise}.
\end{lem}
With Lemma~\ref{lem:bound_noise}, we have
\[
\EB[\|\nabla f(\bx_{t_m}; \xi_{t_m}) - \nabla f(\bx_{t_m}) \|^2|\FM_{t_m}]
\le  C_1  + C_2 \|\bx_{t_m} -\x^*\|^2.
\]
Then, it follows that
\begin{align*}
\EB[\|\sh_m \|^2|\FM_{t_m}]
&\le  \|\nabla f(\bx_{t_m})\|^2 + \frac{C_1}{E_m} + \frac{C_2}{E_m}  \|\bx_{t_m} -\x^*\|^2.
%&\le \frac{C_1}{E_m} + \left(  2L  + \frac{2C_2}{\mu E_m}   \right) \left( f(\bx_{t_m}) - f(\x^*) \right)
\end{align*}
%where we use the $L$-smoothness and $\mu$-strongly convexity of $f(\cdot)$.

\paragraph{Part 2:}
For $\EB[\|\sdelta_m\|^2|\FM_{t_m}]$, by Jensen's inequality, we have
\begin{align*}
\EB[\|\sdelta_m\|^2|\FM_{t_m}]
&=\EB[\|\sv_m - \sh_m\|^2 |\FM_{t_m}] \\
&= \EB \left[\left\|\frac{1}{E_m}  \sum_{t=t_m}^{t_{m+1}-1} \sum_{k=1}^K p_k \nabla f_k(\x_t^k; \xi_t^k)  -
\frac{1}{E_m}  \sum_{t=t_m}^{t_{m+1}-1} \sum_{k=1}^K p_k \nabla f_k(\bx_{t_m}; \xi_t^k) \right\|^2  \bigg|\FM_{t_m} \right]\\
&\le\frac{1}{E_m}  \sum_{t=t_m}^{t_{m+1}-1} \sum_{k=1}^K p_k  \EB \left[\left\|  \nabla f_k(\x_t^k; \xi_t^k)  - \nabla f_k(\bx_{t_m}; \xi_t^k) \right\|^2\big|\FM_{t_m} \right].
\end{align*}
Because $\x_t^k, \bx_{t_m} \in \FM_{t}$ and $\FM_{t_m} \subseteq \FM_{t}$ for $t_m \le t < t_{m+1}$, we have that
\begin{align*}
\EB [\|  \nabla f_k(\x_t^k; \xi_t^k)  - \nabla f_k(\bx_{t_m}; \xi_t^k) \|^2|\FM_{t_m} ]
&=\EB [\EB [\|  \nabla f_k(\x_t^k; \xi_t^k)  - \nabla f_k(\bx_{t_m}; \xi_t^k) \|^2|\FM_{t} ] |\FM_{t_m} ]\\ 
&=\EB [\EB_{\xi_t^k}\|  \nabla f_k(\x_t^k; \xi_t^k)  - \nabla f_k(\bx_{t_m}; \xi_t^k) \|^2 |\FM_{t_m} ]\\ 
&\le L^2 \EB [ \| \x_t^k - \bx_{t_m}\|^2 |\FM_{t_m} ],
\end{align*}
where the first equality follows from the tower rule of conditional expectation and the second inequality follows from the expected $L$-smoothness in Assumption~\ref{asmp:convex}.

Combining the last two results, we have
\[
\EB[\|\sdelta_m\|^2|\FM_{t_m}] 
\le \frac{L^2}{E_m} \sum_{t=t_m}^{t_{m+1}-1} \sum_{k=1}^K p_k \EB [ \| \x_t^k - \bx_{t_m}\|^2 |\FM_{t_m} ]
:=  \frac{L^2}{E_m} \sum_{t=t_m}^{t_{m+1}-1} V_t,
\]
where $V_t$ is the residual error defined by
\begin{equation}
\label{eq:residual_error}
V_t =\sum_{k=1}^K p_k \EB[\| \x_t^k - \bx_{t_m}\|^2|\FM_{t_m}].
\end{equation}
The residual error is incurred by multiple local gradient descents.
Intuitively, if no local update is used (i.e., $E_m=1$), such a residual error would disappear.
The following lemma helps us bound $\frac{1}{E_m} \sum_{t=t_m}^{t_{m+1}-1} V_t$ in terms of $\gamma_m$ and $\|\bx_{t_m}-\x^*\|^2$.
\begin{lem}
	\label{lem:bound_V}
	Under Assumptions~\ref{asmp:convex} and~\ref{asmp:noise}, there exist some universal constants $C_3, C_4 >0$ such that for any $m$ with $\gamma_m^2\frac{E_m-1}{E_m}  C_4 \le 1$, it follows that
	\[
	\frac{1}{E_m} \sum_{t=t_m}^{t_{m+1}-1} V_t 
	\le \gamma_m^2 \frac{E_m-1}{E_m}  \left( C_3 + C_4 \|\bx_{t_m}-\x^*\|^2\right).
	\]
\end{lem}

\paragraph{Almost sure convergence:}
Denote $\Delta_m = f(\bx_{t_m}) - f(\x^*)$ for simplicity, then from the $\mu$-strongly convexity and $L$-smoothness of $f(\cdot)$, it follows that 
\[
\frac{\mu}{2}\|\bx_{t_{m}}-\x^*\|^2 \le \Delta_m \le \frac{1}{2\mu}\|\nabla f(\bx_{t_{m}})\|^2 
\quad \text{and} \quad
\frac{1}{2L}\|\nabla f(\bx_{t_{m}})\|^2  \le \Delta_m \le  \frac{L}{2}\|\bx_{t_{m}}-\x^*\|^2 .
\]
Note that $\gamma_m \to 0$ when $m$ goes to infinity, which means there exists some $m_0$, such that for any $m \ge m_0$, we have $\gamma_m^2 C_4 \le 1$ and $\gamma_m \le \min\{\frac{1}{2L}, 1\}$.
It implies that we can apply Lemma~\ref{lem:bound_V} for sufficiently large $m$.
Combining the two parts and plugging them into~\eqref{eq:iterate_fx} yield for any $m \ge m_0$,
\begin{align}
\label{eq:iterate_Delta_final}
\EB[\Delta_{m+1}|\FM_{t_m}]
&\le \Delta_m  - \frac{\gamma_m}{2}\|\nabla f(\bx_{t_m})\|^2+ \gamma_m^2L \cdot \left[\|\nabla f(\bx_{t_m})\|^2 + \frac{C_1}{E_m} + \frac{C_2}{E_m}  \|\bx_{t_m} -\x^*\|^2\right] 
\nonumber\\ 
& \qquad\qquad +\left(  \frac{\gamma_m}{2} + \gamma_m^2L\right)\gamma_m^2L^2 \left( C_3 + C_4\|\bx_{t_m} -\x^*\|^2  \right) \nonumber \\
&\le \Delta_m  - \gamma_m\mu \Delta_m + \gamma_m^2L \cdot \left[\frac{C_1}{E_m} + \left(  2L  + \frac{2C_2}{\mu E_m}   \right)\Delta_m\right] 
\nonumber\\ 
& \qquad\qquad +\left(  \frac{\gamma_m}{2} + \gamma_m^2L\right)\gamma_m^2 L^2 \left( C_3 + \frac{2C_4}{\mu} \Delta_m\right) \nonumber \\
&\le \Delta_m  - \gamma_m\mu \Delta_m + \gamma_m^2L \cdot \left[C_1+ \left(  2L  + \frac{2C_2}{\mu}   \right)\Delta_m\right] 
+\gamma_m^3L^2 \left( C_3 + \frac{2C_4}{\mu} \Delta_m\right) \nonumber \\
&\le \Delta_m  - \gamma_m\mu \Delta_m + \gamma_m^2L \cdot \left[C_1+ \left(  2L  + \frac{2C_2}{\mu}   \right)\Delta_m\right] 
+\gamma_m^2 L^2\left( C_3 + \frac{2C_4}{\mu} \Delta_m\right) \nonumber \\
&= \left( 1 + c_1 \gamma_m^2  \right)\Delta_m  + c_2 \gamma_m^2 -
\mu\gamma_m \Delta_m, 
\end{align}
where
\[
c_1 = 2L^2 + \frac{2(LC_2+L^2C_4)}{\mu}
\quad \text{and} \quad
c_2 =L C_1 + L^2C_3.
\]
To conclude the proof, we need to apply the Robbins-Siegmund theorem~\citep{robbins1971convergence}.
\begin{lem}[Robbins-Siegmund theorem]
	\label{lem:RS_theorem}
	Let $\{ D_m, \beta_{m}, \alpha_{m}, \zeta_{m}  \}_{m=0}^{\infty}$ be non-negative and adapted to
	a filtration $\{\GM_{m} \}_{m=0}^{\infty}$, satisfying
	\[
	\EB[D_{m+1}|\GM_{m}] \le (1+ \beta_{m}) D_{m} + \alpha_{m} - \zeta_{m}
	\]
	for all $m \ge 0$ and both $\sum_{m} \beta_{m} < \infty$ and $\sum_{m} \alpha_{m} < \infty$ almost surely.
	Then, with probability one, $D_{m}$ converges to a non-negative random variable $D_{\infty} \in [0,   \infty)$ and $\sum_{m} \zeta_{m} < \infty$.
\end{lem}
From Assumption~\ref{asmp:gamma}, we have that $c_1\sum_{m = m_0}^{\infty} \gamma_m^2 < \infty$ and $c_2\sum_{m = m_0}^{\infty} \gamma_m^2 < \infty$.
Hence, based on~\eqref{eq:iterate_Delta_final}, Lemma~\ref{lem:RS_theorem} implies that $\Delta_m = f(\bx_{t_m}) - f(\x^*)$ converges to a finite non-negative random variable $\Delta_{\infty}$ almost surely.
Moreover, Lemma~\ref{lem:RS_theorem} also ensures that
\begin{equation}
\label{eq:finite_gamma_De}
\mu \sum_{m=m_0}^{\infty} \gamma_m \Delta_m < \infty.
\end{equation}
If $\PB(\Delta_m > 0) > 0$, then the left-hand side of~\eqref{eq:finite_gamma_De} would be infinite with positive probability due to the fact $\sum_{m = m_0}^{\infty} \gamma_m = \infty$.
It reveals that $\PB(\Delta_m = 0) =1$ and thus $f(\bx_{t_m}) \to f(\x^*)$ as well as $\bx_{t_m} \to \x^*$ with probability one when $m$ goes to infinity.

\paragraph{$L_2$ convergence:}
We will obtain the $L_2$ convergence rate from~\eqref{eq:iterate_Delta_final}.
This part follows the same argument of~\citet{su2018uncertainty} (see Page 37-38 therein).
For completeness, we conclude this section by presenting the proof of it.
Taking expectation on both sides of~\eqref{eq:iterate_Delta_final},
\[
\frac{ \EB\Delta_{m+1}}{\gamma_{m}}
\le \frac{\gamma_{m-1}\left( 1 - \mu \gamma_m  + c_1 \gamma_m^2  \right)}{\gamma_{m}}\frac{\EB\Delta_m}{\gamma_{m-1}}  + c_2 \gamma_m.
\]
Because $\gamma_m \to 0$, we have that for sufficiently large $m$, $ c_1 \gamma_m^2  \le 0.5 \mu \gamma_m $, and hence, 
\[
\frac{ \EB\Delta_{m+1}}{\gamma_{m}}
\le \frac{\gamma_{m-1}\left( 1 - \frac{\mu}{2} \gamma_m  \right)}{\gamma_{m}}\frac{\EB\Delta_m}{\gamma_{m-1}}  + c_2 \gamma_m.
\]

\begin{lem}[Lemma A.10 in~\citet{su2018uncertainty}]
	Let $c_1, c_2$ be arbitrary positive constants. 
	Assume $ \gamma_m \to 0$ and $\frac{\gamma_{m-1}}{\gamma_{m}} = 1 + o(\gamma_m)$.
	If $B_m > 0$ satisfies $B_m  \le \frac{\gamma_{m-1}\left( 1 - c_1 \gamma_m  \right)}{\gamma_{m}} B_{m-1} + c_2 \gamma_{m}$, then $\sup_{m} B_m < \infty$.
\end{lem}
With the above lemma, we claim that there exists some $C_5 > 0$ such that
\begin{equation}
\label{eq:L2-convergence}
\sup_{0 < m < \infty} \frac{\EB\Delta_m}{\gamma_{m-1}} < C_5,
\end{equation}
which immediately concludes that
\[
\EB\|\bx_{t_m} - \x^*\|^2 \le \frac{2}{\mu} \EB\Delta_m \le \frac{2C_5}{\mu} \gamma_{m-1} =  \frac{2C_5}{\mu} (1+o(\gamma_{m}))\gamma_{m} \le C_0 \gamma_{m}.
\]

\subsection{Proof of Lemma~\ref{lem:bound_noise}}
\begin{proof}
	By Assumption~\ref{asmp:noise}, we know that $\eps(\bx_{t_m}) := \nabla f(\bx_{t_m}; \xi_{t_m})  - \nabla f(\bx_{t_m})$ satisfies
	\[
	\|\EB_{\xi_{t_m}} \eps(\bx_{t_m}) \eps(\bx_{t_m})^\top - \sS\| 
	\le  C\left( \|\bx_{t_m} -\x^*\|  + \|\bx_{t_m} -\x^*\|^2 \right).
	\]
	Therefore, it follows that 
	\begin{align*}
	\label{eq:bound_grad}
	\EB[\|\nabla f(\bx_{t_m}; \xi_{t_m}) - \nabla f(\bx_{t_m}) \|^2|\FM_{t_m}]
	&= \EB [\|\eps(\bx_{t_m})\|^2| \FM_{t_m}] = \EB_{\xi_{t_m}}\|\eps(\bx_{t_m})\|^2  \nonumber \\
	&= \mathrm{tr}(\EB_{\xi_{t_m}} \eps(\bx_{t_m}) \eps(\bx_{t_m})^\top)  \nonumber \\
	&\le d \|\EB_{\xi_{t_m}} \eps(\bx_{t_m}) \eps(\bx_{t_m})^\top\| \nonumber \\
	& \le  d \|\sS\| + dC  \|\bx_{t_m} -\x^*\|+dC  \|\bx_{t_m} -\x^*\|^2 \nonumber \\
	&\le  \left(d \|\sS\| + \frac{dC}{2} \right)+ \frac{3dC}{2}  \|\bx_{t_m} -\x^*\|^2 \nonumber \\
	&\le C_1  + C_2 \|\bx_{t_m} -\x^*\|^2
	\end{align*}
	with $C_1 = d\max_k\|\sS_k\| + \frac{dC}{2}$ and $C_2 = \frac{3dC}{2}$.
	Here we use the fact that $\sS =  \sum_{k=1}^K p_k^2 \sS_k$ and thus $ \|\sS\|\le \sum_{k=1}^K p_k^2 \|\sS_k\| \le \sum_{k=1}^K p_k \|\sS_k\| \le \max_{k \in [K]}\|\sS_k\|$.
	
	With a similar argument, it follows that
	\[
	\EB_{\xi_{t}^k} \|\eps_k(\x_{t}^k)\|^2 \le d\|\sS_k\| + \frac{dC}{2} + \frac{3dC}{2}\|\x_{t}^k-\x^*\|^2\le C_1 + C_2 \|\x_{t}^k-\x^*\|^2.
	\]
\end{proof}

\subsection{Proof of Lemma~\ref{lem:bound_V}}
For a fixed $m \ge 0$, let us consider the case where $t_{m+1} > t_m + 1$, otherwise the result follows directly due to $V_{t_m} = 0$.
For $t_m \le t < t_{m+1}-1$ and $k \in [K]$, we have $\x_{t_m}^k = \bx_{t_m}$ and 
\[
\x_{t+1}^k = \x_t^k - \eta_m \nabla f_k(\x_t^k; \xi_t^k)
\quad  \Rightarrow \quad 
\x_{t+1}^k = \bx_{t_m} -  {\color{red}\eta_m}\sum_{\tau = t_m}^{t} \nabla f_k(\x_{\tau}^k; \xi_{\tau}^k).
\]
%Here we slightly abuse the notation of $\x_{t_{m+1}}^k$ a little bit, which also effects the definition of $V_{t_{m+1}}$ accordingly.
%Local SGD will force $\x_{t_{m+1}}^k$ to be identical among different $k$'s via a global synchronization, while in the above formula, $\x_{t_{m+1}}^k$ doesn't satisfies the requirement.
%One should interpret $\x_{t_{m+1}}^k$ as the non-synchronized updated sequence (only) in this subsection.
%Besides, since our target $\frac{1}{E_m} \sum_{t=t_m}^{t_{m+1}-1} V_t$ has nothing to do with $\x_{t_{m+1}}^k$, the latter only serves as an auxiliary variable. 

Using the last iteration relation, we obtain that
\begin{align*}
\EB[\|\x_{t+1}^k - \bx_{t_m}\|^2 | \FM_{t_m} ]
&=\eta_m^2\EB\left[\left\| \sum_{\tau = t_m}^{t} \nabla f_k(\x_{\tau}^k; \xi_{\tau}^k)\right\|^2 \bigg| \FM_{t_m} \right]\\
&\le \eta_m^2(t+1-t_m)\sum_{\tau = t_m}^{t} \EB[\|  \nabla f_k(\x_{\tau}^k; \xi_{\tau}^k)\|^2 \big| \FM_{t_m} ]\\
&\le \eta_m^2E_m \sum_{\tau = t_m}^{t} \EB[\|  \nabla f_k(\x_{\tau}^k; \xi_{\tau}^k) \|^2 \big| \FM_{t_m} ]\\
&= \eta_m^2E_m \sum_{\tau = t_m}^{t} \EB\left[ \EB(\|  \nabla f_k(\x_{\tau}^k; \xi_{\tau}^k) \|^2\big|  \FM_{\tau}) \big| \FM_{t_m} \right].
\end{align*}
We then turn to bound $\EB[\|  \nabla f_k(\x_{\tau}^k; \xi_{\tau}^k) \|^2 \big| \FM_{\tau} ]$ as  follows:
\begin{align*}
\EB[\|  \nabla f_k(\x_{\tau}^k; \xi_{\tau}^k) \|^2 \big| \FM_{\tau} ]
&= \EB[\|  \nabla f_k(\x_{\tau}^k; \xi_{\tau}^k) -  \nabla f_k(\x_{\tau}^k) \|^2  \big| \FM_{\tau} ]   + \|\nabla f_k(\x_{\tau}^k) \|^2\\
&\le \EB_{\xi_{\tau}^k} \|\eps_k(\x_{\tau}^k)\|^2  + 2\|\nabla f_k(\x_{\tau}^k) - \nabla f_k(\x^*) \|^2+ 2\|\nabla f_k(\x^*) \|^2\\
&\le \left(C_1+ 2\|\nabla f_k(\x^*) \|^2\right)  + \left(C_2+2 L^2\right) \|\x_{\tau}^k-\x^*\|^2  \\
&\le C_3 + \frac{C_4}{2} \|\x_{\tau}^k-\x^*\|^2 \\
&\le  C_3 + C_4 \|\x_{\tau}^k-\bx_{t_m}\|^2 + C_4\|\bx_{t_m} -\x^*\|^2,
\end{align*} 
where $C_3 = C_1 + 2\max_{k \in [K]}\|\nabla f_k(\x^*) \|^2$ and $C_4 = 2C_2+4L^2$.
The second inequality uses the $L$-smoothness to bound $\|\nabla f_k(\x_{\tau}^k) - \nabla f_k(\x^*) \|$ and Lemma~\ref{lem:bound_noise} to bound $\EB_{\xi_{\tau}^k} \|\eps_k(\x_{\tau}^k)\|^2$ which yields 
\[
\EB_{\xi_{\tau}^k} \|\eps_k(\x_{\tau}^k)\|^2
\le  C_1  + C_2\|\x_{\tau}^k-\x^*\|^2.
\]
Therefore, by combing the last two results, we have
\[
\EB[\|\x_{t+1}^k - \bx_{t_m}\|^2 | \FM_{t_m} ]
\le \eta_m^2E_m  \sum_{\tau = t_m}^{t} \left[  C_3 + C_4 \|\bx_{t_m} -\x^*\|^2   
+ C_4\EB[\|\x_{\tau}^k-\bx_{t_m}\|^2| \FM_{t_m} ]
\right].
\]
% For better clarification, for $t_m \le t \le t_{m+1}$, we define an another residual sequence 
%\[
%\widetilde{V}_t = \sum_{k=1}^Kp_k
%\EB(\|\x_{t}^k - \bx_{t_m}\|^2 | \FM_{t_m}). 
%\]
%Recalling the definition of $V_t =  \sum_{k=1}^Kp_k
%\EB(\|\x_{t}^k - \bx_{t_m}\|^2 | \FM_{t_m})$ for $t_m \le t < t_{m+1}$, we can see $\widetilde{V}_t = V_t$ for $t_m \le t < t_{m+1}$ while $V_{t_{m+1}} =0 \le \widetilde{V}_{t_{m+1}}$.
Hence, for $t_m \le t < t_{m+1}-1$, we have
\begin{equation}
\label{eq:residual_1}
V_{t+1}
= \sum_{k=1}^Kp_k
\EB(\|\x_{t+1}^k - \bx_{t_m}\|^2 | \FM_{t_m} ) 
\le \eta_m^2E_m \sum_{\tau = t_m}^{t}\left(  C_3 + C_4 \|\bx_{t_m} -\x^*\|^2   
+ C_4  V_{\tau}  \right).
\end{equation}
Because $V_{t_m} = 0$, it then follows that
\begin{align*}
\frac{1}{E_m} \sum_{t = t_m}^{t_{m+1}-1}  V_{t}  
&=\frac{1}{E_m} \sum_{t=t_m}^{t_{m+1}-2} V_{t+1}\\
&\le  \eta_m^2\sum_{t=t_m}^{t_{m+1}-2} \sum_{\tau = t_m}^{t} \left(  C_3 + C_4 \|\bx_{t_m} -\x^*\|^2   
+ C_4  V_{\tau}  \right)	\\
&= \eta_m^2   \sum_{t = t_m}^{t_{m+1}-2}  (t_{m+1}-t-1) \left(  C_3 + C_4 \|\bx_{t_m} -\x^*\|^2   
+ C_4  V_{t}  \right)	\\
&\le \eta_m^2 (E_m-1)  \sum_{t = t_m}^{t_{m+1}-1}  \left(  C_3 + C_4 \|\bx_{t_m} -\x^*\|^2   
+ C_4  V_{t}  \right)	\\
&\le \gamma_m^2  \frac{E_m-1}{E_m} \left(  C_3 + C_4 \|\bx_{t_m} -\x^*\|^2   
+  \frac{C_4}{E_m} \sum_{t = t_m}^{t_{m+1}-1}    V_{t}  \right),
\end{align*}
where we use the definition of $E_m = t_{m+1}-t_m$ and $\gamma_m = \eta_m E_m$.

Hence, rearranging the last inequality and using the condition $\gamma_m^2  \frac{E_m-1}{E_m} C_4 \le \frac{1}{2}$ gives
\[
\frac{1}{E_m} \sum_{t=t_m}^{t_{m+1}-1} V_t 
\le 2\gamma_m^2 \frac{E_m-1}{E_m}  \left( C_3 + C_4 \|\bx_{t_m} -\x^*\|^2\right).
\]
Finally redefining $C_3:=2C_3$ and $C_4:= 2C_4$ completes the proof and the restriction on $\gamma_m$ becomes $\gamma_m^2\frac{E_m-1}{E_m}  C_4 \le 1$ under the new notation of $C_4$.

\subsection{Proof of Lemma~\ref{lem:T1}}

Recall that 
\[
\seps_m = \sh_m - \nabla f(\bx_{t_{m}}) = \frac{1}{E_m}  \sum_{t=t_m}^{t_{m+1}-1} \left(\nabla f(\bx_{t_m}; \xi_t) - \nabla f(\bx_{t_m})\right)
\]
where $\nabla f(\bx_{t_m}; \xi_t) = \sum_{k=1}^Kp_k\nabla f(\bx_{t_{m}}; \xi_t^k)$ and $\xi_t = \{ \xi_t^k \}_{k \in [K]}$, and
recall that $\eps(\bx_{t_{m}}) = \nabla f(\bx_{t_{m}}; \xi_{t_m}) - \nabla f(\bx_{t_{m}})$.
Hence $\seps_m$ is the mean of $E_m$ i.i.d.\ copies of $\eps(\bx_{t_{m}})$ at a fixed $\bx_{t_{m}}$.

Define $\FM_t = \sigma(\{ \xi_{\tau}^k \}_{1 \le k  \le K,0\le  \tau < t})$ by the natural filtration generated by $\xi_{\tau}^k$'s and $\GM_{m-1} = \FM_{t_m}$.
Then $\{\seps_m\}_{m=1}^{\infty}$ is a martingale difference with respect to $\{\GM_m\}_{m=0}^{\infty}$ (for convention $\GM_0 = \{ \emptyset, \Omega \}$ if $\bx_0$ is deterministic, otherwise $\GM_0 = \sigma(\bx_0)$): 
$\EB[\seps_m|\GM_{m-1}] = \0$.

The following lemma establishes an invariance principle which allows us to extend traditional martingale CLT.
Interesting readers can find its proof in~\citet{hall2014martingale} (see Theorems 4.1, 4.2 and 4.4 therein).
\begin{lem}[Invariance principles in the martingale CLT]
	\label{lem:L_martingal_CLT}
Let $\{S_n, \GM_n\}_{n \ge 1}$ be a zero-mean, square-integrable martingale with difference $X_n = S_n - S_{n-1} (S_0 = 0)$.
Let $U_n^2 = \sum_{m=1}^n \EB[X_m^2|\GM_{m-1}]$ and $s_n^2 = \EB U_n^2 = \EB S_n^2$.
Define $\zeta_n(t)$ as the linear interpolation among the points $(0, 0)$, $(U_n^{-2}U_1^2, U_n^{-1}S_1)$, $(U_n^{-2}U_2^2, U_n^{-1}S_2)$, $\dots$, $(1, U_n^{-1}S_n)$, namely, for $t \in [0 ,1]$ and $0 \le i \le n-1$,
\[
\zeta_n(t) : = U_n^{-1} \left[ S_i + (U_{i+1}^2-U_i^2)^{-1}(tU_n^2 - U_i^2) X_{i+1} \right]
\quad \text{if} \quad
U_i^2 \le t U_n^2 <U_{i+1}^2.
\]
As $n \to \infty$, if (i) the Linderberg conditions holds, namely for any $\eps > 0$,
\begin{equation}
\label{eq:Lindeberg}
s_n^{-2} \sum_{m=1}^n \EB[X_m^2I(|X_m| \ge \eps s_n)] \to 0,
\end{equation}
and (ii) $s_n^{-2} U_n^2 \to 1$ almost surely and $s_n^2 \to \infty$, then
\[
\zeta_n(t) \Rightarrow B(t) 
\quad \text{in the sense of} \quad (C, \rho).
\]
Here $B(t)$ is the standard Brownian motion on $[0, 1]$ and $C = C[0, 1]$ is the space of real, continuous functions on $[0, 1]$ with the uniform metric $\rho: C[0, 1] \to [0, \infty)$, $\rho(\omega) = \max_{t \in [0, 1]} |\omega(t)|$.
\end{lem}

Lemma~\ref{lem:L_martingal_CLT} is for univariate martingales. 
We will use the {Cram\'{e}r-Wold device} to reduce the issue of convergence of multivariate martingales to univariate ones.
To that end, we fix any uni-norm vector $\a$ and define $X_m = \a^\top\seps_m$.
We then check the two conditions in Lemma~\ref{lem:L_martingal_CLT} hold for such $\{X_m, \GM_m\}_{m \ge 1}$.

\paragraph{The Linderberg condition:}
For one thing, since $\bx_{t_m} \to \x^*$ almost surely from Lemma~\ref{lem:a.s.convergence}, we have $\EB\| \eps(\bx_{t_m})\|^{2+\delta_2} \precsim 1$ from Assumption~\ref{asmp:noise} when $m$ is sufficiently large.
\begin{lem}[Marcinkiewicz–Zygmund inequality and Burkholder inequality]
\label{lem:MZ}
	If $Z_1, \dots, Z_n$ are independent random vectors such that $\EB Z_m = 0$ and $\EB |Z_m|^p < \infty$ for $1 \le p < \infty$, then
	\[
	\EB\left| \frac{1}{n}\sum_{m=1}^n Z_m \right|^p
	\le \frac{C_p}{n^{\frac{p}{2}}}\EB \left( \frac{1}{n}\sum_{m=1}^n \left| Z_m \right|^2\right)^\frac{p}{2},
	\]
	where the $C_p$ are positive constants which depend only on $p$ and not on the underlying distribution of the random variables involved.
	If $Z_1, \dots, Z_n$ are martingale difference sequence, the above inequality still holds. 
	It is named as Burkholder's inequality~\citep{dharmadhikari1968bounds}.
\end{lem}
 Notice that we can rewrite $X_m$ as the mean of $E_m$ i.i.d.\ random variables which have the same distribution as $Z_1 = \a^\top\eps(\bx_{t_{m}})$: $X_m  = \frac{1}{E_m}\sum_{i=1}^{E_m} Z_i$.
With the Marcinkiewicz–Zygmund inequality and Jensen inequality, it follows that 
\begin{align}
\label{eq:varaince_em}
\EB|X_m|^{2+\delta_2} 
&\precsim  E_m^{-(1+\frac{\delta_2}{2})}  \EB\left(\frac{1}{n}\sum_{m=1}^n \left| Z_m \right|^2 \right)^{1+\frac{\delta_2}{2}} 
\precsim  E_m^{-(1+\frac{\delta_2}{2})}  \EB\left| Z_1 \right|^{2+\delta_2}\nonumber \\
&\precsim  E_m^{-(1+\frac{\delta_2}{2})}  \|\a\|^{2+\delta_2}
\EB\|\eps(\bx_{t_{m}})\|^{2+\delta_2}
\precsim E_m^{-1}.
\end{align}
Moreover, from Assumption~\ref{asmp:noise} and Lemma~\ref{lem:a.s.convergence},  we have that
\begin{align*}
\left|{\a^\top\left[\EB \eps(\bx_{t_{m}})\eps(\bx_{t_{m}})^\top - \sS\right]\a }\right|
&\le C\left[ \EB\|\bx_{t_{m}}-x^*\| + \EB\|\bx_{t_{m}}-x^*\|^2 \right]\\
&\le  C ( \sqrt{\gamma_m} + \gamma_m) \to 0.
\end{align*}
Recall that $\sum_{m=1}^T  {E_m}^{-1} \to \infty$ as $T \to \infty$.
The Stolz–Cesàro theorem (Lemma~\ref{lem:SC}) implies that
\begin{equation}
\label{eq:s_T}
\lim_{T \to \infty} \frac{s_T^2 }{\sum_{m=1}^T  \frac{1}{E_m} \a^\top\sS \a}
=\lim_{T \to \infty}\frac{\sum_{m=1}^T  \frac{\a^\top\EB \eps(\bx_{t_{m}})\eps(\bx_{t_{m}})^\top \a}{E_m} }{\sum_{m=1}^T  \frac{1}{E_m} \a^\top\sS \a}
=\lim_{T \to \infty} \frac{ \a^\top\EB \eps(\bx_{t_{T}})\eps(\bx_{t_{T}})^\top \a}{\a^\top\sS \a}
= 1.
\end{equation}
Hence, for any $\eps > 0$, as $T \to \infty$, we have that
\begin{align*}
s_T^{-2} \sum_{m=1}^T \EB[X_m^{2}I(|X_m| \ge \eps s_T)]
&\le \eps^{-{\delta_2}}s_T^{-(2+{\delta_2})} \sum_{m=1}^T \EB[|X_m|^{2+{\delta_2}}I(|X_m| \ge \eps s_T)]\\
&\le \eps^{-{\delta_2}}s_T^{-(2+{\delta_2})} \sum_{m=1}^T \EB|X_m|^{2+{\delta_2}}\\
&\precsim \eps^{-{\delta_2}}s_T^{-(2+{\delta_2})} \sum_{m=1}^T
\frac{1}{E_m}\\
& \asymp \eps^{-{\delta_2}}s_T^{-{\delta_2}} \to 0.
\end{align*}

\paragraph{The second condition:}
We have established the divergence of $\{\s_T^2\}_{T}$ in~\eqref{eq:s_T}.
Notice that
\begin{align*}
U_T^2 &= \sum_{m=1}^T \EB[X_m^2|\GM_{m-1}] 
= \sum_{m=1}^T \frac{1}{E_m} \a^\top\EB[\eps(\bx_{t_m})\eps(\bx_{t_m})^\top |\GM_{m-1}]\a\\
&= \sum_{m=1}^T \frac{1}{E_m} \a^\top\EB_{\xi_{t_m}}\eps(\bx_{t_m})\eps(\bx_{t_m})^\top\a.
\end{align*}
Therefore, from~\eqref{eq:s_T} and the Stolz–Cesàro theorem (Lemma~\ref{lem:SC}), it follows almost surely that
\begin{align*}
\lim_{T \to \infty}\left| \frac{U_T^2}{s_T^2} -1  \right|
&\le \lim_{T \to \infty} \frac{C}{s_T^2} \sum_{m=1}^T\frac{1}{E_m} \left[ \|\bx_{t_{m}}-\x^*\| +\|\bx_{t_{m}}-\x^*\|^2 \right]\\
&= \lim_{T \to \infty} \frac{C}{\a^\top \sS \a} \left[ \|\bx_{t_{T}}-\x^*\| +\|\bx_{t_{T}}-\x^*\|^2 \right] \to 0.
\end{align*}

\begin{lem}[Stolz–Cesàro theorem]
	\label{lem:SC}
Let $\{a_n\}_{n \ge 1}$ and  $\{b_n\}_{n \ge 1}$ be two sequences of real numbers.
Assume that $\{b_n\}_{n \ge 1}$  is a strictly monotone and divergent sequence.
We have that
\[
\text{if} \ \lim_{n \to \infty} \frac{a_{n+1} -a_{n}}{b_{n+1}-b_n} = l, \
\text{then} \
\lim_{n \to \infty} \frac{a_{n}}{b_n} = l.
\]
\end{lem}

We have shown that the two conditions in Lemma~\ref{lem:L_martingal_CLT} hold.
Hence, by definition, $\zeta_T(r)  \Rightarrow B(r) $ where
\[ 
\zeta_T(r) : = U_T^{-1} \left[ S_i + (U_{i+1}^2-U_i^2)^{-1}(rU_T^2 - U_i^2) X_{i+1} \right]
\quad \text{if} \quad
U_i^2 \le r U_T^2 <U_{i+1}^2
\]
and $S_i = \sum_{m=1}^i X_m$.
Since $s_T/U_T \to 1$ almost surely and~\eqref{eq:s_T}, it follows that
\[
\frac{\sqrt{t_T}}{T}U_T\zeta_T(r) \Rightarrow  \sqrt{\nu} \sqrt{\a^\top\sS\a} B(r) \overset{d.}{=} \sqrt{\nu}\a^\top\sS^{1/2} \B_d(r),
\]
where $\B_d(r)$ is the $d$-dimensional standard Brownian motion.
Recall that
\[
h(r, T) = \max\left\{ n \in \ZB_+ \bigg|  r \sum_{m=1}^T \frac{1}{E_m}  \ge  \sum_{m=1}^n \frac{1}{E_m}  \right\}.
\]
\begin{lem}
\label{lem:uniform_small}
Under the same condition of Lemma~\ref{lem:T1}, it follows that
\[
\sup_{r \in [0, 1]} \left| \frac{\sqrt{t_T}}{T}U_T\zeta_T\left(\frac{U_{h(r, T)}^2}{U_T^2} \right) -\frac{\sqrt{t_T}}{T}U_T\zeta_T\left(r \right) \right| \to 0
\quad \text{in probability}.
\]
\end{lem}
Hence,
\begin{align*}
\frac{\sqrt{t_{T}}}{T} \sum_{m=1}^{h(r, T)}\a^\top\seps_m
=\frac{\sqrt{t_{T}}}{T} S_{h(r, T)}
=\frac{\sqrt{t_T}}{T}U_T\zeta_T\left(\frac{U_{h(r, T)}^2}{U_T^2} \right) 
\Rightarrow   \sqrt{\nu}\a^\top\sS^{1/2} \B_d(r).
\end{align*}
By the arbitrariness of  $\a$, it follows that\footnote{See the proof of Theorem 4.3.5. in~\citet{whitt2002stochastic} for more detail about how to argue multivariate weak convergence from univariate weak convergence along any direction.} 
\[
	\frac{\sqrt{t_{T}}}{T} \sum_{m=1}^{h(r, T)} \seps_m \Rightarrow  \sqrt{\nu} \sS^{1/2} \B_d(r) .
\]
Applying the continuous mapping theorem to the linear function $\seps: \seps \mapsto \sG^{-1}\seps$, we have
\[
	\frac{\sqrt{t_{T}}}{T} \sum_{m=1}^{h(r, T)}  \sG^{-1}\seps_m \Rightarrow  \sqrt{\nu}  \sG^{-1}\sS^{1/2} \B_d(r).
\]
Finally, since $\EB \frac{\sqrt{t_{T}}}{T} \| \sG^{-1}\seps_0\| \to 0$, it implies that $\frac{\sqrt{t_{T}}}{T}\sG^{-1}\seps_0 = o_{\PB}(1)$.
Then it is clear that $\frac{\sqrt{t_{T}}}{T} \sum_{m=0}^{h(r, T)}  \sG^{-1}\seps_m \Rightarrow  \sqrt{\nu}  \sG^{-1}\sS^{1/2} \B_d(r).$

\subsection{Proof of Lemma~\ref{lem:uniform_small}}
From the Theorem A.2 of~\citet{hall2014martingale}, if some random function $\phi_n \Rightarrow \phi$ in the sense of $(C, \rho)$, $\{\phi_n\}$ must be tight in the sense that for any $\eps > 0$, $\PB(\sup_{|s-t|\le \delta}|\phi_n(s)-\phi_n(t)|\ge \eps) \to 0$ uniformly in $n$ as $\delta \to 0$.
Since $\frac{\sqrt{t_T}}{T}U_T\zeta_T(r) \Rightarrow  \sqrt{\nu}\a^\top\sS^{1/2} \B_d(r)$,  $\{\frac{\sqrt{t_T}}{T}U_T\zeta_T\}_T$ is tight.
We denote the following notation for simplicity
\[
\phi_T(r) = \frac{\sqrt{t_T}}{T}U_T\zeta_T(r)
\quad \text{and} \quad
p_T(r) = \frac{U_{h(r, T)}^2}{U_T^2}.
\]

Since $p_T(r)$ satisfies $p_T(0) = 1 - p_T(1) = 0$ and $p_T(r)$ is non-decreasing and right-continuous in $r$, we can view $p_T(r)$ as the cumulative distribution function of some random variable on $[0, 1]$ and $p(r): r \mapsto r$ is the cumulative distribution function of uniform distribution on $[0, 1]$.
It is clearly that $p_T(r) \to p(r)$ for every $r \in [0, 1]$ almost surely, because
\begin{align*}
\lim_{T \to \infty}   p_T(r)
= \lim_{T \to \infty}  \frac{U_{h(r, T)}^2}{U_T^2} 
= \lim_{T \to \infty} \frac{s_{h(r, T)}^2}{s_T^2}
= \lim_{T \to \infty}\frac{\sum_{m=1}^{h(r, T)} \frac{1}{E_m}}{\sum_{m=1}^T \frac{1}{E_m}} = r = p(r).
\end{align*}
Here we use $h(r, T) \to \infty$ for any $r \in [0, 1]$ as $T \to \infty$.
 Since $p(\cdot)$ is additionally continuous, weak convergence implies uniform convergence in cumulative distribution functions, i.e.,
\begin{equation}
\label{eq:uniform_r}
\lim_{T \to \infty} \sup_{r \in [0, 1]}\left| p_T(r) -r \right| = 0.
\end{equation}

By the tightness of $\{\phi_n\}$, for any $\eps, \eta> 0$, we can find a sufficiently small $\delta$ such that 
\[
\limsup_{T \to \infty}\PB\left(\sup_{|s-t|\le \delta}|\phi_T(s)-\phi_T(t)|\ge \eps\right) \le \eta.
\]
With~\eqref{eq:uniform_r}, for this $\delta$, $\PB( \sup_{r \in [0, 1]}|p_T(r)  - r| > \delta ) \to 0$ as $T \to \infty$.
It implies that
\begin{eqnarray*}
\lefteqn{ \limsup_{T \to \infty} \PB\left(\sup_{r \in [0, 1]}  |\phi_T\left( p_T(r) \right) -\phi_T(r)| \ge \eps\right) } \\
&\le & \limsup_{T \to \infty} \PB\left( \sup_{r \in [0, 1]}  |\phi_T\left( p_T(r) \right) -\phi_T(r)| \ge \eps, \sup_{r \in [0, 1]}|p_T(r)  - r| \le \delta \right) \\
& & \qquad  + \lim_{T \to \infty} \PB\left(\sup_{r \in [0, 1]} |p_T(r)  - r| > \delta \right) \\
 &\le & \limsup_{T \to \infty} \PB\left(\sup_{|s-t|\le \delta}|\phi_T(s)-\phi_T(t)|\ge \eps\right) 
 \le \eta.
\end{eqnarray*}
Because $\eta$ is arbitrary, we have shown that
\[
\sup_{r \in [0, 1]}  |\phi_T\left( p_T(r) \right) -\phi_T(r)|  \to 0 
\quad \text{in probability}.
\]

\subsection{Proof of Lemma~\ref{lem:T2}}

Recall that $\sG = \nabla^2 f(\x^*), \s_m = \bx_{t_m}-\x^*$ and
\[
\sr_m = \nabla f(\bx_{t_m}) - \sG \s_m.
\]
When $\|\s_m\| \le \delta_1$, by Assumption~\ref{asmp:convex}, $\| \nabla^2 f(s \s_m + \x^*) - \nabla^2 f(\x^*)\| \le s L' \|\s_m\|$, then
\begin{align*}
\|\sr_m\| 
&= \| \nabla f(\s_m + \x^*) - \nabla f(\x^*)- \nabla^2 f(\x^*)  \s_m \| \\
&=\left\| \int_{0}^1 \nabla^2 f(s \s_m + \x^*)\s_m d s - \nabla^2 f(\x^*)  \s_m \right\| \\
&\le \int_{0}^1 \left\| \nabla^2 f(s \s_m + \x^*) - \nabla^2 f(\x^*)\right\|\|\s_m\|ds \\
&\le \frac{L'}{2} \|\s_m\|^2.
\end{align*}
When $\|\s_m\| > \delta_1$, $\|\sr_m\|  \le \|\nabla f(\bx_{t_{m}})\| + \|\sG \s_m\| \le L \|\s_m\| +  L \|\s_m\| = 2 L \|\s_m\|$.
Applying the results above yields
\[
\|\sr_m\|  \le L' \|\s_m\|^2 1_{\{\|\s_m\| \le \delta_1\}} + 2 L \|\s_m\|1_{\{\|\s_m\| > \delta_1\}}.
\]
Hence,
\begin{align*}
\frac{\sqrt{t_{T}}}{T} \sum_{m=0}^T \left\| \sr_m\right\| 
&\le  \frac{\sqrt{t_{T}}}{T} \sum_{m=0}^T \left[L' \|\s_m\|^2 1_{\{\|\s_m\| \le \delta_1\}} + 2 L \|\s_m\|1_{\{\|\s_m\| > \delta_1\}} \right].
\end{align*}
By Lemma~\ref{lem:a.s.convergence}, $\s_m \to 0$ almost surely, which implies
\[
\frac{\sqrt{t_{T}}}{T}\sum_{m=0}^T  \|\s_m\|1_{\{\|\s_m\| > \delta_1\}} \to 0 
\quad \text{almost surely}.
\]
It then suffices to show that $ \frac{\sqrt{t_{T}}}{T} \sum_{m=0}^T  \|\s_m\|^2 1_{\{\|\s_m\| \le \delta_1\}} = o_{\PB}(1)$, which is implied by 
\[
\frac{\sqrt{t_{T}}}{T} \sum_{m=0}^T  \EB \|\s_m\|^2 = o(1).
\]
It holds because $\frac{\sqrt{t_{T}}}{T} \sum_{m=0}^T  \EB \|\s_m\|^2 \precsim \frac{\sqrt{t_{T}}}{T} \sum_{m=0}^T \gamma_m \to 0$ from Lemma~\ref{lem:a.s.convergence} and Assumption~\ref{asmp:E}.

\subsection{Proof of Lemma~\ref{lem:T3}}
In the proof of Lemma~\ref{lem:a.s.convergence} (see the Part 2 therein), we have established for sufficiently large $m$,
\[
\EB[\|\sdelta_m\|^2|\FM_{t_m}]
\le \frac{L^2}{E_m} \sum_{t=t_m}^{t_{m+1}-1} V_t
\le L^2\gamma_m^2 \frac{E_m-1}{E_m}  \left( C_3 + C_4 \|\bx_{t_m}-\x^*\|^2\right),
\]
where $V_t$ is the residual error defined in~\eqref{eq:residual_error} and $C_3, C_4 > 0$ are universal constants defined in Lemma~\ref{lem:bound_V}.
Besides, Lemma~\ref{lem:a.s.convergence} implies that $\EB\|\bx_{t_m}-\x^*\|^2 \precsim \gamma_m \precsim 1$.
It follows that
\[
\EB \|\sdelta_m\|^2 
\le L^2\gamma_m^2 \left( C_3 + C_4 \EB\|\bx_{t_m}-\x^*\|^2\right)
\precsim \gamma_m^2.
\]
In order to prove the conclusion, it suffices to show that $ \frac{\sqrt{t_{T}}}{T} \sum_{m=0}^T  \EB \left\|\sdelta_m\right\| \to 0$, 
which is satisfied because
\[
\frac{\sqrt{t_{T}}}{T} \sum_{m=0}^T  \EB \left\|\sdelta_m\right\|
\le \frac{\sqrt{t_{T}}}{T} \sum_{m=0}^T  \sqrt{\EB \left\|\sdelta_m\right\|^2}
\precsim  \frac{\sqrt{t_{T}}}{T} \sum_{m=0}^T \gamma_m \to 0
\]
from Lemma~\ref{lem:a.s.convergence} and Assumption~\ref{asmp:E}.

\subsection{Proof of Lemma~\ref{lem:iterate}}
If $\{E_m\}$ is uniformly bounded (i.e., there exists some $C$ such that $1 \le E_m \le C$ for all $m$), the conclusion follows because
\[
0 \le  \frac{(\sum_{m=0}^{T-1} E_m)(\sum_{m=0}^{T-1} E_m^{-1}a_{m, T})}{T^2}  
\le  \frac{CT(\sum_{m=0}^{T-1} a_{m, T})}{T^2}  = \frac{1}{T}\sum_{m=0}^{T-1} a_{m, T} \to 0 \quad \text{when} \quad T \to \infty.
\]
In the following, we instead assume $E_m$ is non-decreasing in $m$ (i.e., $1\le E_{m} \le E_{m+1}$ for all $m$).
Let $H_k = \sum_{m=0}^k a_{m, T}$. 
For any $\varepsilon$, there exist some $N = N(\varepsilon)$, such that for any $m \ge N$, $0 \le H_m \le m \varepsilon$.
Then
\begin{align*}
\sum_{n=N}^T \frac{a_{m, T}}{E_m}
&= \sum_{n=N}^T \frac{H_m-H_{m-1}}{E_m}
=   \frac{H_T}{E_T} + \sum_{n=N}^{T-1}  \left(\frac{1}{E_m}-\frac{1}{E_{m+1}}\right)H_m - \frac{H_{N-1}}{E_N}\\
&\le   \frac{H_T}{E_T} + \sum_{n=N}^{T-1}  \left(\frac{1}{E_m}-\frac{1}{E_{m+1}}\right)m\varepsilon - \frac{H_{N-1}}{E_N}\\
&=\frac{H_T-T\varepsilon}{E_T} + \left[ \frac{T\varepsilon}{E_T} + \sum_{n=N}^{T-1}  \left(\frac{1}{E_m}-\frac{1}{E_{m+1}}\right)m\varepsilon - \frac{(N-1)\varepsilon}{E_N} \right]- \frac{H_{N-1}-(N-1)\varepsilon}{E_N}\\
&= \varepsilon \cdot \sum_{n=N}^{T} \frac{1}{E_m} +  \frac{H_T-T\varepsilon}{E_T} - \frac{H_{N-1}-(N-1)\varepsilon}{E_N}\\
&\le \varepsilon \cdot \sum_{n=N}^{T} \frac{1}{E_m} + \frac{N\varepsilon}{E_N}
\end{align*}
Recall $t_T = \sum_{m=0}^{T-1} E_m$. Therefore,
\begin{align*}
\frac{t_T(\sum_{m=0}^{T-1} E_m^{-1}a_{m, T})}{T^2} 
=& \frac{t_T(\sum_{m=0}^{N-1} E_m^{-1}a_{m, T})}{T^2}   + \frac{t_T(\sum_{m=N}^{T-1} E_m^{-1}a_{m, T})}{T^2}  \\
\le& \frac{t_T(\sum_{m=0}^{N-1} E_m^{-1}a_{m, T})}{T^2}  + \varepsilon \frac{t_T(\sum_{m=N}^{T} E_m^{-1})}{T^2}  + \frac{t_TN\varepsilon}{T^2E_N}.
\end{align*}
Taking superior limit on both sides and noting $a_{m, T} \precsim 1$ uniformly and $\lim \limits_{T \to \infty}\frac{t_T}{T^2} = 0$, we have 
\[
0 \le \limsup\limits_{T \to \infty} \frac{t_T(\sum_{m=0}^{T-1} E_m^{-1}a_{m, T})}{T^2} 
\le \varepsilon \nu.
\]
By the arbitrariness of $\varepsilon$, we complete the proof.

\subsection{Proof of Lemma~\ref{lem:sup_epsi}}
Without loss of generality, we assume $\sG^{-1}$ is a positive diagonal  matrix.
Otherwise, we apply the spectrum decomposition to $\sG = \sV \D \sV^\top$ and focus on the coordinates of each $\seps_m$ with respect to the orthogonal base $\sV$.
This simplification reduces our multivariate case to a univariate one.
Hence, it is enough to show that the result holds for one-dimensional $\seps_m$ and $\sG$.
In the following argument, we focus on an eigenvalue $\lambda$ of $\sG$ and its eigenvector $\sv$, and denote $\eps_m = \sv^\top\seps_m$ and $B_m = 1 - \gamma_m \lambda \in \RB$ for simplicity. 
Clearly, $\lambda \ge 0$ and $0 < B_m \le 1$ for sufficiently large $m$.

Given a positive integer $n$, we separate the time interval $[0, T]$ uniformly into $n$ portions with $h_i=\left[\frac{iT}{n}\right] (i=0,1,\dots, n)$ the $i$-th endpoint.
The choice of $n$ is independent of $T$, which implies that $\lim_{T \to \infty} h_i = \infty$ for any $i$.
Define an event $\mathcal{A}$ whose complement is 
\[\mathcal{A}^c=\left\{\exists h_i\text{ s.t. }\left\|\frac{\sqrt{t_T}}{T\gamma_{h_i+1}}\sum_{m=0}^{h_i}  \left(\prod\limits_{i=m+1}^{h_i}B_i\right) \gamma_m  \eps_m\right\|\ge \eps\right\}.
\]
We claim that $\limsup\limits_{T \to \infty} \PB(\AM^c) = 0$.
Indeed, by the union bound and Markov's inequality,
\begin{align*}
\mathbb{P}(\mathcal{A}^c)&\le \sum_{i=0}^{n}\mathbb{P}\left\{\left\|\frac{\sqrt{t_T}}{T\gamma_{h_i+1}}\sum_{m=0}^{h_i}\prod_{j=m+1}^{h_i}B_j \gamma_m\eps_m\right\|\ge \eps\right\}\\
&\precsim \sum_{i=0}^{n}\frac{t_T}{\eps^2T^2\gamma_{h_i+1}^2}\sum_{m=0}^{h_i}\left(\prod_{j={m+1}}^{h_i}B_j\right)^2\gamma_m^2\\
&\precsim \frac{t_T}{\eps^2T^2}\sum_{i=0}^{n} \frac{1}{\gamma_{h_i+1}}\\
&\le \frac{t_T(n+1)}{\eps^2T^2 \gamma_{T+1}} \to 0 
\quad \text{as} \quad T\to \infty.
\end{align*}
Here the last two inequality uses for any $i \in [n]$,
\[
\frac{1}{\gamma_{h_i+1}}\sum_{m=0}^{h_i}\left(\prod_{j={m+1}}^{h_i}B_j\right)^2\gamma_m^2 \precsim 1,
\]
which is implied by
\begin{align*}
&\lim_{h_i\to \infty}\left\{\sum_{m=0}^{h_i}\gamma_m^2\left(\prod_{j=0}^m B_j\right)^{-2}\right\}
\bigg/\left\{\gamma_{h_i}\left(\prod_{j=0}^{h_i}B_j\right)^{-2}\right\}\\
&= \lim_{h_i\to \infty}\left\{\gamma_{h_i}^2\left(\prod_{j=0}^{h_i}B_j^{-2}\right)\right\}
\bigg/\left\{o(\gamma_{h_i-1})\gamma_{h_i-1}\prod_{j=0}^{h_i}B_j^{-2}+\gamma_{h_i}\prod_{j=0}^{h_i}B_j^{-2}(1- B_{h_i}^2)\right\}\\
&= \lim_{h_i\to \infty}\frac{\gamma_{h_i}^2}{o(1)\gamma_{h_i-1}^2+ 2\lambda\gamma_{h_i}^2 - \lambda^2\gamma_{h_i}^3 } \\
&= \frac{1}{2\lambda} < \infty
\end{align*}
as a result of the Stolz–Cesàro theorem (Lemma~\ref{lem:SC}).
Here we observe that the denominator $\gamma_{h_i}\left(\prod_{j=0}^{h_i}B_j\right)^{-2}$ increases in $h_i$ and diverges when $h_i$ is sufficiently large.

Since the event $\mathcal{A}^c$ has diminishing probability, we focus on the event $\mathcal{A}$.
We will prove that on the event $\mathcal{A}$ our target random sequence is uniformly tight.
For notation simplicity, we define
\[
X_m^h = \prod_{i=m}^hB_i.
\]
It follows that
\begin{align*}
&\mathbb{P}\left\{\frac{\sqrt{t_T}}{T}\sup_{0\le h\le T}\left|\frac{1}{\gamma_{h+1}}\sum_{m=0}^h\left(\prod_{i=m+1}^h B_i\right)\gamma_m\eps_m\right|\ge 2\eps\text{ ; }\mathcal{A}\right\}\\
&= \mathbb{P}\left\{\frac{\sqrt{t_T}}{T}\sup_{0\le h\le T}\left|\frac{1}{\gamma_{h+1}X_{h+1}^T}\sum_{m=0}^hX_{m+1}^T\gamma_m\eps_m\right|\ge 2\eps\text{ ; }\mathcal{A}\right\}\\
& \le \sum_{i=0}^{n-1}\mathbb{P}\left\{\frac{\sqrt{t_T}}{T}\sup_{h\in[h_i,h_{i+1})}\left|\frac{1}{\gamma_{h+1}X_{h+1}^T}\left(\sum_{m=0}^h X_{m+1}^T\gamma_m\eps_m\right)\right|\ge 2\eps\text{ ; }\mathcal{A}\right\}\\
&\le  \sum_{i=0}^{n-1}\mathbb{P}\left\{\frac{\sqrt{t_T}}{T}\sup_{h\in[h_i,h_{i+1})}\frac{1}{\gamma_{h+1}X_{h+1}^T}\left|\sum_{m=0}^{h_i}X_{m+1}^T\gamma_m\eps_m+\sum_{m=h_i+1}^hX_m^T\gamma_m\eps_m\right|\ge 2\eps\text{ ; }\mathcal{A}\right\}\\
&\le \sum_{i=0}^{n-1}\mathbb{P}\left\{\frac{\sqrt{t_T}}{T}\sup_{h\in[h_i,h_{i+1})}\left[\frac{1}{\gamma_{h+1}X_{h+1}^T}\left|\sum_{m=0}^{h_i}X_{m+1}^T\gamma_m\eps_m\right|
+\left|\frac{1}{\gamma_hX_{h+1}^T}\sum_{m=h_i+1}^hX_m^T\gamma_m\eps_m\right|\right]\ge 2\eps\text{ ; }\mathcal{A}\right\}\\
&\le \sum_{i=0}^{n-1}\mathbb{P}\left\{\frac{\sqrt{t_T}}{T}\sup_{h\in[h_i,h_{i+1})}\left|\frac{1}{\gamma_{h+1}X_{h+1}^T}\sum_{m=h_i+1}^hX_{m+1}^T\gamma_m\eps_m\right|\ge \eps\text{ ; }\mathcal{A}\right\}\\
&\le \sum_{i=0}^{n-1}\mathbb{P}\left\{\frac{\sqrt{t_T}}{T}\sup_{h\in[h_i,h_{i+1})}\left|\frac{1}{\gamma_{h+1}X_{h+1}^T}\sum_{m={h_i+1}}^hX_{m+1}^T\gamma_m\eps_m\right|\ge \eps\right\}\\
&=  \sum_{i=0}^{n-1}\mathbb{P}\left\{\left(\frac{\sqrt{t_T}}{T}\right)^{2+\delta}\sup_{h\in[h_i,h_{i+1})}\left(\frac{1}{\gamma_{h+1}X_{h+1}^T}\right)^{2+\delta}\left|\sum_{m={h_i+1}}^hX_{m+1}^T\gamma_m\eps_m\right|^{2+\delta}\ge \eps^{2+\delta}\right\}\\
&:= \sum_{i=0}^{n-1}\mathcal{P}_i,
\end{align*}
where $\delta$ is any positive real number less than $\min\{\delta_2, \delta_3\}$.

Let $Y_h=\left|\sum_{m=h_i+1}^hX_{m+1}^T\gamma_m\eps_m\right|^{2+\delta}$.
It is clear that $Y_h$ is a sub-martingale adapted to the natural filtration.
Let $c_h=\frac{1}{(\gamma_hX_h^T)^{2+\delta}}$. Then $\{c_h\}$ is a non-increasing sequence when $h$ is sufficiently large because
\[
\gamma_{h}X_{h}^T=\frac{\gamma_{h}}{\gamma_{h+1}}(1-\lambda\gamma_{h})\gamma_{h+1}X_{h+1}^T=(1+o(\gamma_h))(1-\lambda\gamma_h)\gamma_{h+1}X_{h+1}^T\le \gamma_{h+1}X_{h+1}^T
\]
for sufficiently large $h$.
Indeed, since $h \ge h_i=\left[\frac{iT}{n}\right] \to \infty$ as $T \to \infty$, $(1+o(\gamma_h))(1-\lambda\gamma_h) \le 1$ is solid and $X_h^T$ is non-negative when $T$ goes to infinity.
Hence, each $\mathcal{P}_i$ is the probability of the event where the maximum of a  sub-martingale multiplied by a non-increasing sequence is larger than a threshold.
To bound each $\mathcal{P}_i$, we use Chow's inequality which is a generalization of Doob's inequality~\citep{chow1960martingale}.
It follows that
\begin{align}
\label{eq:bound_P}
\mathcal{P}_i
&= \mathbb{P}\left\{\frac{t_T^{1+\delta/2}}{T^{2+\delta}}\sup_{h\in[h_i,h_{i+1})}c_hY_h\ge \eps^{2+\delta}\right\}\nonumber \\
&\le \frac{t_T^{1+\delta/2}}{\eps^{2+\delta}T^{2+\delta}}\left\{c_{h_{i+1}-1}\mathbb{E}Y_{h_{i+1}-1}+\sum_{j=h_i+1}^{h_{i+1}-2}(c_i-c_{i+1})\mathbb{E}Y_j\right\}.
\end{align}
We then apply  Burkholder's inequality to bound each $\EB Y_j$.
Burkholder's inequality is a generalization of the Marcinkiewicz–Zygmund inequality (Lemma~\ref{lem:MZ}) to martingale differences~\citep{dharmadhikari1968bounds}. That is, 
\begin{align*}
\mathbb{E}Y_j
&=\mathbb{E}\left|\sum_{m=h_i+1}^jX_{m+1}^T\gamma_m\eps_m\right|^{2+\delta}\\
&\precsim (j-h_i)^{\delta/2}\sum_{m=h_i+1}^{j}\EB\left|X_{m+1}^T\gamma_m\eps_m\right|^{2+\delta} \\
&\precsim (j-h_i)^{\delta/2}\sum_{m=h_i+1}^{j}(X_{m+1}^T\gamma_m)^{2+\delta}/E_m^{1+\delta/2}\\
&\precsim (j-h_i)^{\delta/2}\sum_{m=h_i+1}^j c_m^{-1}/E_m^{1+\delta/2},
\end{align*}
where we use $\EB\left|\eps_m\right|^{2+\delta} \precsim  1/E_m^{1+\delta/2}$ for sufficiently large $m$ that is already derived in~\eqref{eq:varaince_em}.

Plugging it into~\eqref{eq:bound_P} yields that $\mathcal{P}_i$ is bounded by
\begin{align*}
% \mathcal{P}_i
&\frac{t_T^{1+\delta/2}}{\eps^{2+\delta}T^{2+\delta}}\left\{c_{h_{i+1}-1}\mathbb{E}Y_{h_{i+1}-1}+\sum_{j=h_i+1}^{h_{i+1}-2}(c_i-c_{i+1})\mathbb{E}Y_j\right\}\\
& \precsim \frac{t_T^{1+\delta/2}}{\eps^{2+\delta}T^{2+\delta}}\left\{c_{h_{i+1}{-}1}(h_{i+1} {-} h_i)^{\frac{\delta}{2}}\sum_{m=h_i+1}^{h_{i+1} {-} 1}\frac{c_m^{-1}}{E_m^{1+\delta/2}}+\sum_{j=h_i+1}^{h_{i+1} {-} 2}(c_j {-} c_{j+1})(j {-} h_i)^{\frac{\delta}{2}}\sum_{m=h_i {+} 1}^{j}\frac{c_m^{-1}}{E_m^{1+\delta/2}}\right\}\\
&\le \frac{t_T^{1+\delta/2}}{\eps^{2+\delta}T^{2+\delta}}\left(\frac{T}{n}\right)^{\delta/2}\left\{c_{h_{i+1}-1}\sum_{m=h_i+1}^{h_{i+1}-1}\frac{c_m^{-1}}{E_m^{1+\delta/2}}+\sum_{j=h_i+1}^{h_{i+1}-2}(c_j-c_{j+1})\sum_{m=h_i+1}^j\frac{c_m^{-1}}{E_m^{1+\delta/2}}\right\}\\
&= \frac{t_T^{1+\delta/2}}{\eps^{2+\delta}T^{2+\delta}}\left(\frac{T}{n}\right)^{\delta/2}\left\{c_{h_{i+1}-1}\sum_{m=h_i+1}^{h_{i+1}-1}\frac{c_m^{-1}}{E_m^{1+\delta/2}}+\sum_{m=h_i+1}^{h_{i+1}-2}(c_m-c_{h_{i+1}-1})\frac{c_m^{-1}}{E_m^{1+\delta/2}}\right\}\\
&= \frac{t_T^{1+\delta/2}}{\eps^{2+\delta}T^{2+\delta}}\left(\frac{T}{n}\right)^{\delta/2}\left\{\sum_{m=h_i+1}^{h_{i+1}-1}c_m\frac{c_m^{-1}}{E_m^{1+\delta/2}}\right\}\\
&= \frac{t_T^{1+\delta/2}}{\eps^{2+\delta}T^{2+\delta}}\left(\frac{T}{n}\right)^{\delta/2}\sum_{m=h_i+1}^{h_{i+1}-1}\frac{1}{E_m^{1+\delta/2}}.
\end{align*}
Recall $t_T=\sum_{m=0}^{T-1} E_m$.
Summing the last bound over $i = 0, 1, \dots, n-1$ gives
\begin{align*}
\sum_{i=0}^{n-1}\mathcal{P}_i
&\precsim \frac{t_T^{1+\delta/2}}{\eps^{2+\delta}T^{2+\delta}}\left(\frac{T}{n}\right)^{\delta/2}\sum_{m=0}^{T-1}\frac{1}{E_m^{1+\delta/2}}\\
&=
\frac{1}{\eps^{2+\delta} n^{\delta/2}}
\frac{(\frac{1}{T}\sum_{m=0}^{T-1}E_m)^{1+\delta/2}}{\frac{1}{T}\sum_{m=0}^{T-1}E_m^{1+\delta/2}}
\frac{\sum_{m=0}^{T-1}E_m^{1+\delta/2}\sum_{m=0}^{T-1}1/E_m^{1+\delta/2}}{T^2}
\\
&\precsim \frac{1}{n^{\delta/2}},
\end{align*}
where we use $(ii)$ in Assumption~\ref{asmp:E} which implies
\[\sup_{T}
\frac{\sum_{m=0}^{T-1}E_m^{1+\delta/2}\sum_{m=0}^{T-1}1/E_m^{1+\delta/2}}{T^2} 
\le \sup_{T} \frac{\sum_{m=0}^{T-1}E_m^{1+\delta_3}\sum_{m=0}^{T-1}1/E_m^{1+\delta_3}}{T^2}
< \infty
\]
as a result of $\delta < \delta_3$. 

Summing them all, we have
\begin{align*}
&\limsup_{T \to \infty}\mathbb{P}\left\{\frac{\sqrt{t_T}}{T}\sup_{0\le h\le T}\left|\frac{1}{\gamma_{h+1}}\sum_{m=0}^h\left(\prod_{i=m+1}^h B_i\right)\gamma_m\eps_m\right|\ge 2\eps\right\}\\
&\le \limsup_{T \to \infty}\mathbb{P}\left\{\frac{\sqrt{t_T}}{T}\sup_{0\le h\le T}\left|\frac{1}{\gamma_{h+1}}\sum_{m=0}^h\left(\prod_{i=m+1}^h B_i\right)\gamma_m\eps_m\right|\ge 2\eps\text{ ; }\mathcal{A}\right\} +\limsup_{T \to \infty} \PB(\AM^c)\\
&\le \limsup_{T \to \infty}\sum_{i=0}^{n-1}\mathcal{P}_i\\
&\precsim \frac{1}{n^{\delta/2}}.
\end{align*}
Since the probability of the left hand side has nothing to do with $n$, letting $n \to \infty$ concludes the proof.

%\begin{align*}
%    \lim_{T\to \infty}&\frac{T^{\delta/2}\sum_{m=0}^{T-1}1/E_m^{1+\delta/2}}{\tilde{t}_T^{1+\delta/2}}\simeq \lim_{T\to \infty}\frac{T^{\delta/2-1}\sum_{m=0}^{T-1}1/E_m^{1+\delta/2}+T^{\delta/2}/E_T^{1+\delta/2}}{\tilde{t}_T^{\delta/2}/E_T}\\
%    & \precsim \lim_{T\to \infty}\left(\frac{T}{\tilde{t}_TE_T}\right)^{\delta/2}+\lim_{T\to \infty}\frac{T^{\delta/2-1}\tilde{t}_T/E_m^{\delta/2}}{\tilde{t}_T^{\delta/2}/E_T}\\
%    &= \lim_{T\to \infty}\left(\frac{T}{\tilde{t}_TE_T}\right)^{\delta/2}+\lim_{T\to \infty}\left(\frac{T}{\tilde{t}_TE_T}\right)^{1-\delta/2}
%\end{align*}
%
%Actually we have
%\begin{align*}
%    \lim_{T\to \infty}\frac{\tilde{t}_TE_T}{T}=\lim_{T\to \infty}\frac{1/E_T}{1/E_T+T\Delta 1/E_T}=\lim_{T\to \infty}\frac{1}{1+T\Delta E_T/E_T}=\Theta(1)
%\end{align*}

%%%%%%%%%%%%%%%%%%%%%%%%%%%%%%%%%%%%%%%%%%%%%%%%%%%%%%%%%%%%%%%%%%%%%%%%%%%%%%%%%%%%%%%%%%%%%%%%%%%
%%%%%%%%%%%%%%%%%%%%%%%%%%%%%%%%%%%%%%%%%%%%%%%%%%%%%%%%%%%%%%%%%%%%%%%%%%%%%%%%%%%%%%%%%%%%%%%%%%%
\section{Proofs of Proposition~\ref{prop:limit-Em}}
\label{appen:limit-Em}

To prove the proposition, we make two following claims.
\paragraph{Claim 1:} For any positive sequences $\{a_n\}$ and $\{b_n\}$ with $\sum\limits_{n=1}^T b_n\rightarrow \infty$, we have 
\begin{equation}
\label{eq:suplim_le}
\limsup\limits_{T\to \infty} \frac{\sum_{n=1}^Ta_n}{\sum_{n=1}^Tb_n}\le \limsup\limits_{T\to \infty}\frac{a_T}{b_T}.
\end{equation}
Without loss of generality, we assume the right hand side is finite, otherwise~\eqref{eq:suplim_le} follows obviously.
We denote that $\limsup\limits_{T\to \infty}\frac{a_T}{b_T}=\lambda$ for simplicity. 
Based on the definition of limit superior, for any $\eps>0$, there exists $N_\eps$ subject to $a_n<(\lambda+\eps)b_n$ for $\forall n\ge N_\eps$. 
As a result, 
\[
\sum\limits_{n=1}^T a_n=\sum\limits_{n=1}^{N_\eps}a_n+\sum\limits_{n=N_\eps+1}^Ta_n\le \sum\limits_{n=1}^{N_\eps}a_n+ (\lambda+\eps)\sum\limits_{n=N_\eps+1}^Tb_n,
\]
which implies
\[
\frac{\sum_{n=1}^Ta_n}{\sum_{n=1}^Tb_n} \le \frac{\sum_{n=1}^{N_\eps}a_n+ (\lambda+\eps)\sum_{n=N_\eps+1}^Tb_n}{\sum_{n=1}^Tb_n}.
\]
Taking limit superior on both sides and noting that $\sum\limits_{n=1}^T b_n\rightarrow \infty$, we have $\frac{\sum_{n=1}^Ta_n}{\sum_{n=1}^Tb_n}\le \lambda+2\eps$.
By the arbitrariness of $\eps$,~\eqref{eq:suplim_le} follows.

\paragraph{Claim 2:} For any non-decreasing sequence $\{E_m\}$ satisfying $\limsup\limits_{T\to \infty} T (1-\frac{E_{T-1}}{E_T}) < 1$, we can find $\delta>0$ such that
\[
T\left(\frac{1}{E_T}\right)^{1+\delta}-(T-1)\left(\frac{1}{E_{T-1}}\right)^{1+\delta}>0.
\]
In fact, we can choose any $\delta < 1 {-} \limsup\limits_{T\to \infty}T(1 {-} \frac{E_{T-1}}{E_T})$.
In this way, for sufficiently large $T$, we have
\begin{align*}
    T\left(\frac{1}{E_T}\right)^{1+\delta}-(T-1)\left(\frac{1}{E_{T-1}}\right)^{1+\delta}
    &=\left(\frac{1}{E_{T-1}}\right)^{1+\delta}\left(T\left(\frac{E_{T-1}}{E_T}\right)^{1+\delta}-T+1\right)\\
    &\ge T\left(\frac{1}{E_{T-1}}\right)^{1+\delta}\left[\left(1-\frac{1-\delta}{T}\right)^{1+\delta}-1+\frac{1}{T}\right].
\end{align*}
To lower bound the right hand side, we consider the auxiliary function $h(x)=(1-(1-\delta)x)^{1+\delta}+x$ where $x\in(0,1)$.
We claim that $h(x) > 1$ for any $x \in (0, 1)$.
We check it by investigating the derivative of $h(\cdot)$,
\[
\dot{h}(x)=-(1+\delta)(1-(1-\delta)x)^{1+\delta}(1-\delta)+1>-(1+\delta)(1-\delta)+1=\delta^2>0.
\]
Therefore, by mean value theorem, $h(x)>h(0)=1$ which proves the claim.

Now we are well prepared to prove the proposition.
It follows that
\begin{align*}
\limsup\limits_{T\to \infty}T\left[1-\left(\frac{E_{T-1}}{E_T}\right)^{1+\delta}\right]
&=\limsup\limits_{T\to\infty}T\frac{(1+\delta)(\theta_T E_T+(1-\theta_T)E_{T-1})^\delta(E_{T}-E_{T-1})}{E_{T}^{1+\delta}}\\
&\le (1+\delta)\limsup\limits_{T\to \infty}\left(\frac{\theta_T E_T+(1-\theta_T)E_{T-1}}{E_T}\right)^\delta\limsup\limits_{T\to \infty}T\frac{E_T-E_{T-1}}{E_T}\\
&\le (1+\delta)(1-\delta)\limsup\limits_{T\to \infty}\left(\frac{\theta_T E_T+(1-\theta_T)E_{T-1}}{E_T}\right)^\delta\\
&\le 1-\delta^2,
\end{align*}
where the first equality uses mean value theorem with some $\theta_{T} \in [0, 1]$.

Therefore,
\begin{eqnarray*}
    \lefteqn{\limsup\limits_{T\to \infty}\frac{(\sum_{m=1}^T E_m^{1+\delta})(\sum_{m=1}^T (1/E_m)^{1+\delta})}{T^2} } \\
    &\overset{(a)}{\le} & \limsup\limits_{T\to \infty}\frac{E_T^{1+\delta}\sum_{m=1}^T(1/E_m)^{1+\delta}+(\sum_{m=1}^T E_m^{1+\delta})/(E_T)^{1+\delta}}{2T-1}\\
    &\le & \limsup\limits_{T\to \infty}\frac{\sum_{m=1}^T(1/E_m)^{1+\delta}}{(2T-1)/E_T^{1+\delta}}+\frac{1}{2} \\
    &< & \limsup\limits_{T\to \infty}\frac{\sum_{m=1}^T(1/E_m)^{1+\delta}}{T(1/E_T)^{1+\delta}}\\
    &\overset{(b)}{\le} & \limsup\limits_{T\to\infty}\frac{(1/E_T)^{1+\delta}}{T(1/E_T)^{1+\delta}-(T-1)(1/E_{T-1})^{1+\delta}}\\
    &\le & \limsup\limits_{T\to\infty}\frac{1}{1-T\left[1-\left(\frac{E_{T-1}}{E_T}\right)^{1+\delta}\right]}\\
    &\le & \left\{1-\limsup\limits_{T\to\infty}T\left[1-\left(\frac{E_{T-1}}{E_T}\right)^{1+\delta}\right]\right\}^{-1}\le \delta^{-2}<\infty,
\end{eqnarray*}
where (a) uses Claim 1 and (b) uses Claim 1 and Claim 2 together.

Furthermore, if the sequence $\{E_m\}$ satisfies $\lim\limits_{T\to\infty}T\left(1-\frac{E_{T-1}}{E_T}\right)=\rho < 1$, then by the Stolz–Cesàro theorem (Lemma \ref{lem:SC}), we have 
\begin{eqnarray*}
   \lefteqn{ \lim\limits_{T\to\infty}\frac{(\sum_{m=1}^TE_m)(\sum_{m=1}^T1/E_m)}{T^2} }  \\
    &=& \lim\limits_{T\to\infty}\frac{E_T(\sum_{n=1}^T1/E_n)+(\sum_{n=1}^{T-1}E_n)/E_T}{2T-1}\\
    &=& \frac{1}{2} \left\{\lim\limits_{T\to\infty}\frac{\sum_{n=1}^T1/E_n}{T/E_T}+\lim\limits_{T\to\infty}\frac{\sum_{n=1}^TE_n}{TE_T}\right\} \\
    &=& \frac{1}{2} \left\{ \lim\limits_{T\to\infty}\frac{1/E_T}{T/E_T-(T-1)/E_{T-1}}+\lim\limits_{T\to\infty}\frac{E_T}{TE_T-(T-1)E_{T-1}}\right\}  \\
    &=&  \frac{1}{2} \left\{ \lim\limits_{T\to\infty}\frac{E_{T-1}}{E_T}\times\frac{1}{1-T(1-E_{T-1}/E_T)}+\lim\limits_{T\to\infty}\frac{1}{1+(T-1)(1-E_{T-1}/E_T)}\right\} \\
    &=& \frac{1}{2} \left\{\frac{1}{1-\rho}+\frac{1}{1+\rho}\right\}  = \frac{1}{1-\rho^2},
\end{eqnarray*}
which completes the proof.

%%%%%%%%%%%%%%%%%%%%%%%%%%
%%%%%%%%%%%%%%%%%%%%%%%%%%%%%%%%%%%%%%%%%%%%%%%%%%%%%%%%%%%%%%%%%%%%%%%%%%%%%%%%%%%%%%%%%%%%%%%%%%
\section{Proof for the Plug-in Method, Theorem~\ref{thm:G-and-S}}

For simplicity, we denote $\nabla f(\x; \xi_t) = \sum_{k=1}^K p_k \nabla f_k(\x; \xi_t^k)$ and $\nabla^2 f(\x; \xi_t) = \sum_{k=1}^K p_k \nabla^2 f_k(\x; \xi_t^k)$ where $\xi_t = \{\xi_t^k\}_{k \in [K]}$.
We decompose $\widehat{\sG}_T - \sG$ into the following terms:
\begin{align}
\label{eq:G-decom}
\widehat{\sG}_T - \sG
&=\frac{1}{T}\sum_{m=1}^T\nabla^2 f(\bx_{t_m}; \xi_{t_m}) - \sG \nonumber \\
&=\left[\frac{1}{T}\sum_{m=1}^T\nabla^2 f(\x^*; \xi_{t_m}) - \sG\right]
+ \frac{1}{T}\sum_{m=1}^T\left[\nabla^2 f(\bx_{t_m}; \xi_{t_m}) -\nabla^2 f(\x^*; \xi_{t_m})\right].
\end{align}
The first term in~\eqref{eq:G-decom} is asymptotically zero due to the strong law of large number.
With Theorem~\ref{thm:clt}, we have known that under the condition, $\EB\|\bx_{t_m}-\x^*\| \le \sqrt{\EB\|\bx_{t_m}-\x^*\|^2} \lesssim \sqrt{\gamma_m}$.
Then the second term in~\eqref{eq:G-decom} can be bounded via Assumption~\ref{asmp:H}
\begin{align*}
\EB\left\|  \frac{1}{T}\sum_{m=1}^T\left[\nabla^2 f(\bx_{t_m}; \xi_{t_m}) -\nabla^2 f(\x^*; \xi_{t_m})\right] \right\|
&\le \frac{1}{T}\sum_{m=1}^T \EB\left\|  \nabla^2 f(\bx_{t_m}; \xi_{t_m}) -\nabla^2 f(\x^*; \xi_{t_m})\right\| \\
&\le\frac{L'' }{T}\sum_{m=1}^T \EB\left\|\bx_{t_m}-\x^*\right\| \\
&\lesssim \frac{1}{T}\sum_{m=1}^T\sqrt{\gamma_m} \to 0
\end{align*}
as $T \to \infty$.
Hence, $\widehat{\sG}_T$ converges to $\sG$ in probability.

For $\widehat{\sS}_T$, note that
\[
\nabla f(\bx_{t_m}; \xi_{t_m}) 
= \nabla f(\x^*; \xi_{t_m}) + \left[ \nabla f(\bx_{t_m}; \xi_{t_m}) - \nabla f(\x^*; \xi_{t_m})\right] := \sC_m  + \sD_m.
\]
We decompose $\widehat{\sS}_T - \sS$ into the following terms:
\[
\widehat{\sS}_T -\sS = \left( \frac{1}{T}\sum\limits_{m=1}^{T} \sC_m \sC_m^\top - \sS \right)
+ \frac{1}{T}\sum\limits_{m=1}^{T} \sC_m \sD_m^\top
+ \frac{1}{T}\sum\limits_{m=1}^{T} \sD_m \sC_m^\top
+ \frac{1}{T}\sum\limits_{m=1}^{T} \sD_m \sD_m^\top.
\]
Because $\{\sC_m\}_m$ are i.i.d.\ and $\EB\sC_m\sC_m^\top = \sS$, the first term is asymptotically zero due to the strong law of large number.
Note that $\EB\| \sC_m \|^2 = \EB\| \sC_m \sC_m^\top \| \le \mathrm{tr}(\EB\sC_m \sC_m^\top) = \mathrm{tr}(\sS)$ and 
\begin{align*}
\EB \|\sD_m\|^2 
&= \EB  \left\|\sum_{k=1}^K p_k \left(\nabla f_k(\bx_{t_m}; \xi_{t_m}^k) - \nabla f(\x^*; \xi_{t_m}^k)\right)\right\|^2\\
&\le\sum_{k=1}^k p_k \EB  \left\|\nabla f_k(\bx_{t_m}; \xi_{t_m}^k) - \nabla f(\x^*; \xi_{t_m}^k)\right\|^2\\
&\le L^2 \EB\|\bx_{t_m} - \x^*\|^2 \lesssim \gamma_m.
\end{align*}
Then, the second and third terms can be bounded via
\begin{align*}
\EB\left\| \frac{1}{T}\sum\limits_{m=1}^{T} \sC_m \sD_m^\top \right\|
&\le \frac{1}{T}\sum\limits_{m=1}^{T} \EB\| \sC_m \| \| \sD_m\|\\
&\le \frac{1}{T}\sum\limits_{m=1}^{T} \sqrt{\EB\| \sC_m \|^2 \EB \| \sD_m\|^2}\\
&\lesssim \frac{1}{T}\sum\limits_{m=1}^{T} \sqrt{\gamma_m} \to 0.
\end{align*}
Finally, for the last term, we have that
\begin{align*}
\EB \left\| \frac{1}{T}\sum\limits_{m=1}^{T} \sD_m \sD_m^\top \right\|
\le\frac{1}{T}\sum\limits_{m=1}^{T} \EB \left\|  \sD_m \right\|^2 
\lesssim \frac{1}{T}\sum\limits_{m=1}^{T} \gamma_m \to 0.
\end{align*}
Hence, $\widehat{\sS}_T$ converges to $\sS$ in probability.

\end{appendix}

\end{document}